\documentclass{article}

\usepackage[preprint]{corl_2026}
\makeatletter
\renewcommand{\@noticestring}{$^*$Core contributors.}
\makeatother

\usepackage{booktabs}
\usepackage{graphicx} 
\usepackage{amsmath}
\usepackage{amssymb}
\usepackage{comment}
\usepackage{wrapfig}
\usepackage{float}
\usepackage[table, dvipsnames]{xcolor}
\usepackage{hyperref}
\usepackage{tabularx}
\usepackage{pifont}
\usepackage{siunitx}
\usepackage{pgfplots}
\usepackage{pdflscape}
\usepackage{rotating}
\usepackage{longtable}
\usepackage{censor}
\usepackage{arydshln}

\newcommand{\rcs}{RCS}
\newcommand{\vla}{VLA}

\newcommand{\ballmaze}{Ball-Maze}
\newcommand{\binsort}{Bin-Sort}
\newcommand{\blockbalance}{Block-Balance}
\newcommand{\carrypot}{Carry-Pot}
\newcommand{\hingechest}{Hinge-Chest}
\newcommand{\joinblocks}{Join-Blocks}
\newcommand{\pourmarbles}{Pour-Marbles}
\newcommand{\springdoor}{Spring-Door}
\newcommand{\transfercube}{Transfer-Cube}
\newcommand{\transfergate}{Transfer-Gate}
\newcommand{\transferreorient}{Transfer-Reorient}
\newcommand{\asupp}{Asymmetric Support}
\newcommand{\bimani}{Bimanual Manipulation}
\newcommand{\sequhand}{Sequential Handoff}
\newcommand{\paraexe}{Parallel Execution}

\newcommand{\franka}{FR3}
\newcommand{\frankaduo}{FR3 Duo}
\newcommand{\duobench}{DuoBench}

\newcommand{\piofive}{\texorpdfstring{$\pi_{0.5}$}{Pi 0.5}}
\newcommand{\act}{ACT}
\newcommand{\xvla}{X-VLA}
\newcommand{\passk}{\rho_k}

\title{DuoBench: A Reproducible Benchmark for Bimanual Manipulation in Simulation and the Real World}

\author{
    Tobias J\"ulg$^{1*}$, Seongjin Bien$^{1*}$, Simon Hilber$^2$, Yannik Blei$^1$, Pierre Krack$^1$,\\
    \textbf{Maximilian Li$^2$, Sven Parusel$^3$, Rudolf Lioutikov$^2$, Florian Walter$^4$, Wolfram Burgard$^1$}\\
    $^1$University of Technology Nuremberg, $^2$Karlsruhe Institute of Technology\\
    $^3$Franka Robotics, $^4$Technical University of Munich
}

\begin{document}
\maketitle

\begin{abstract}
    Bimanual robot systems substantially expand manipulation capabilities, but coordinating two arms introduces additional control complexity and failure modes that are not well captured by existing benchmarks. We introduce DuoBench, an extensible benchmarking framework for bimanual manipulation policies on the \frankaduo{} platform. DuoBench comprises eleven tasks spanning four coordination categories, implemented in simulation and partially replicated in the real world through reproducible task recipes with 3D-printable assets. In addition, we propose a stage-based evaluation scheme that supports fine-grained semantic failure analysis beyond binary success. It includes stage-conditional pass rates with confidence intervals as well as human-teleoperated datasets for all benchmark tasks. We benchmark several imitation-learning and vision-language-action policies in simulation and on real hardware. Our results show that current policies remain challenged by bimanual manipulation, particularly at the beginning of interaction, during parallel arm execution, and in sim-to-real transfer settings. We further find that simulation reveals the same failure modes as real-world experiments. DuoBench provides a reproducible testbed for analyzing them and studying future methods for dual-arm policy learning.
    Code, datasets, and videos are available at 
    \href{https://duobench.github.io/}{https://duobench.github.io}
\end{abstract}


\begin{figure}[H]
	\centering
	\includegraphics[width=\linewidth]{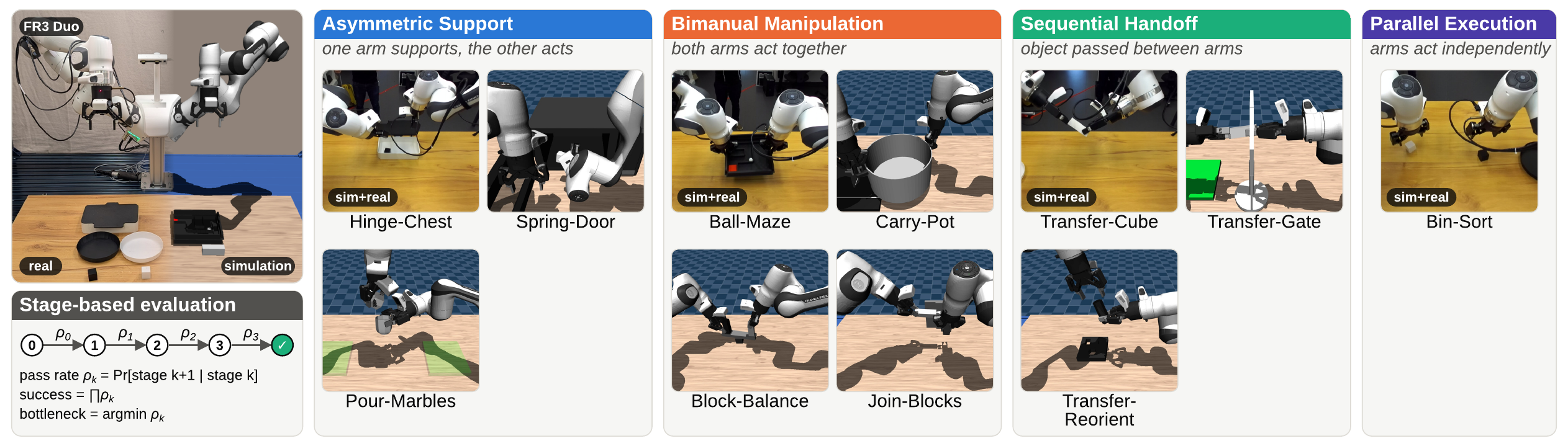}
	\caption{Overview of DuoBench: four bimanual task categories with eleven tasks, four of which are replicated in the real world (left: real-world setup and simulated MuJoCo scene). Each task is decomposed into task stages to better understand policy failure modes. We provide a sim-to-real teleoperation pipeline to facilitate data collection and support extensibility.}
	\label{fig:overview}
\end{figure}

\section{Introduction}
\label{sec:intro}

The relevance of bimanual manipulation is increasing with the growing use of dual-arm tabletop systems and humanoid robots. Many manipulation problems fundamentally require two coordinated arms, yet benchmark development has focused much more heavily on single-arm settings. This gap is particularly limiting for long-horizon bimanual tasks, where binary success rates provide only coarse feedback and fail to reveal which type of coordination causes a policy to break down. At the same time, task reproducibility in real-world deployment settings remains a central challenge. Rather than assuming direct generalization from one domain to the other, a benchmark should make tasks reproducible across both settings and provide users with the tools to collect their own teleoperated data. This is increasingly viable with new robot hardware such as the \frankaduo{} platform, which packages two standard \franka{} arms into a human-inspired dual-arm setup and is likely to become more accessible to labs that already use \franka{} hardware. In summary, a systematic benchmark for bimanual Vision-Language-Action (\vla{}) model manipulation should combine reproducible sim-and-real tasks with fine-grained failure analysis in order to expose model weaknesses and support more targeted progress.

Despite recent progress in manipulation benchmarking, existing benchmarks still provide only limited support for the systematic evaluation of bimanual coordination. In particular, current settings often underrepresent the diversity of coordination patterns required by two-arm manipulation, provide little support for reproducible sim-and-real evaluation, or rely primarily on binary task success. As a result, they offer only limited insight into whether failures arise from grasp acquisition, arm coordination, object transfer, or later task execution.

To address this gap, we introduce DuoBench, a benchmark for bimanual manipulation on the \frankaduo{} platform that combines simulation, reproducible real-world task recipes, and human-teleoperated datasets. DuoBench spans eleven tasks across four coordination categories and augments task success rates with stage-based evaluation, enabling the fine-grained analysis of semantic failure modes. In addition, the benchmark is designed to support low-effort data collection and future extensions with new tasks through a shared teleoperation and simulation interface.

Our main contributions are:
    (1) We introduce DuoBench, a reproducible benchmarking framework for bimanual manipulation with eleven tasks
    in simulation and partially in the real world.
    (2) We propose a taxonomy that organizes bimanual manipulation tasks into four distinct categories.
    (3) We introduce task stages as a fine-grained evaluation mechanism for semantic failure analysis beyond binary success, reported with stage-conditional pass rates and confidence intervals in simulation and on the real robot.
    (4) We provide human-teleoperated datasets together with reproducible real-world task recipes for benchmarking dual-arm policies across simulation and physical setups.

\section{Related Work}
\label{sec:related_work}

With the increasing popularity of machine learning-based approaches for robot control, the demand for methods to compare the performance of new algorithms and for efficient data generation has risen in recent years. Numerous benchmarks have been proposed to meet this role, often built on top of existing open-source robot simulation software such as MuJoCo~\cite{mujoco}, IsaacSim~\cite{isaacsim}, or SAPIEN~\cite{sapien}.

\paragraph{Single-Arm Manipulation in Simulation}
Earlier benchmarks like RLBench~\cite{rlbench} and \mbox{MetaWorld}~\cite{metaworld} were specifically designed for reinforcement learning, while more recent works focus on imitation learning and often include procedurally generated or teleoperated datasets. The ManiSkill~\cite{maniskill,maniskill2,maniskill3} benchmark series features a wide variety of robots and tasks, some of which are bimanual, and its latest version supports fast GPU-based simulation. The LIBERO benchmark~\cite{libero}, built on top of robosuite~\cite{robosuite}, is widely used for benchmarking \vla{}s. Recent extensions~\cite{liberoplus,liberopro,liberox} introduce perturbations along different dimensions to make the benchmark more realistic and meaningful. Another line of work focuses on household scenarios~\cite{robocasa, robocasa365} or addresses long-horizon tasks~\cite{ikea, calvin}. VLABench~\cite{vlabench} also targets long-horizon tasks but specifically addresses the benchmarking of \vla{} by including tasks requiring world knowledge. VIMA~\cite{vima} focuses on multimodal prompting. The diversity of benchmarks has also motivated the development of frameworks providing unified interfaces~\cite{robohive, roboverse}.

\paragraph{Dual-Arm Manipulation in Simulation}
Most relevant to our work are benchmarks designed for dual-arm manipulation. RLBench2~\cite{rlbench2} features 13 tasks to be completed by two table-mounted Franka robots facing each other, but the majority of their tasks only test tightly coupled motions. RoboTwin~\cite{robotwin2} consists of 16 tasks, but they can be interpreted as object-type and episode-length ablations across six basic task categories.
BiCoord~\cite{bicoord} introduces 18 tasks that maximize simultaneous operation of both arms with a focus on long horizon, but many of their tasks are ablations on object types, instead of being unique coordination challenges. RoboEval~\cite{roboeval} evaluates eight bimanual tasks in simulation with stagewise progress and behavioral metrics for efficiency, coordination, and safety, and is closest to our stage-based evaluation. BEHAVIOR~\cite{behavior} provides 100 household activities along with a formal language to generate task instances. As goals are specified by logic clauses, the number of satisfied goal literals can be seen as a measure of the degree to which a task is completed. The task stages introduced in our work also provide a metric for task progress. Bi-DexHands~\cite{bidexhands} defines 20 tasks based on the Fine Motor Subtest. ST-BiBench~\cite{stbibench} investigates coordination and defines classes for parallel and collaborative manipulation. The tasks of DuoBench are defined following a bimanual task taxonomy that also includes tasks from these categories. Benchmarks for humanoid robots also include bimanual manipulation tasks~\cite{humanoidbench, bigym}.

\paragraph{Real-World and Hybrid Benchmarks}
There are comparably few real-world benchmarks, which can be explained by the effort required to design reproducible hardware experiments. The most common method to achieve reproducibility is to use 3D-printed objects. FMB~\cite{fmb} provides 66 procedurally generated object types and focuses on compositing basic skills to solve long-horizon tasks. The objects in FurnitureBench~\cite{heo2023furniturebench} are inspired by IKEA furniture. Importantly, like our work, this benchmark also includes a corresponding simulation environment. However, both the simulation and the real-world setup use a single robot. RoboMind~\cite{robominddataset} also includes bimanual tasks, but provides only the final dataset without the setup for reproducing it. Finally, RoboArena~\cite{roboarena} aims to address reproducibility issues in real-world benchmarks by proposing a method for crowd-sourcing pairwise policy evaluations using the DROID dataset~\cite{droiddataset}.

What is still missing is a bimanual manipulation benchmark with a systematic selection of tasks that supports reproducible evaluation in simulation and in the real world. DuoBench addresses this gap with tasks motivated by a bimanual task taxonomy, implemented both as simulation environments and as recipes for real-world setups. It further provides the software stack for data collection, inference, and seamless switching between simulation and physical experiments.  Our tasks emphasize coordination and motion type variety, requiring semantic-level understanding of the actions rather than repetitions of similar motions across object-level variations for generalization.

\section{Methodology}
\label{sec:method}

The benchmark is built on top of the Robot~Control~Stack~(RCS)~\cite{rcs} ecosystem. As such, all benchmark environments in both simulation and real-world deployments are composed of a base Markov Decision Process (MDP)~\cite{sutton1998reinforcement} $M=\langle S, A, P, R\rangle$ with state space $S$, action space $A$, state transition probability distribution $P$ and reward distribution $R$. The environment is wrapped by $n$ wrappers $W = \langle f: S \to S^\prime, g: A^\prime \to A, P^\prime, R^\prime\rangle$ where each wrapper creates a new wrapped environment: $M^\prime = W \rhd M = \langle S^\prime, A^\prime, P^\prime, R^\prime \rangle$.

A stochastic agent is defined as $\pi(a\mid s) = P_\pi(A_t=a \mid S_t=s)$. In our context, an agent can either be a human teleoperator or a learned policy $\pi_\theta$.
A data recorder wrapper $W_r$ stores the action $a_t\in\mathbb{R}^{16}$ (seven joint and one gripper dimension for each arm) and the delayed observation $s_{t+1} = (I, p_{t+1}, C_{1,t+1}, \ldots , C_{n,t+1})$ where $I$ is the task instruction, $p_t\in\mathbb{R}^{16}$ the proprioception and $C_{i,t+1}$ is the frame from the i-th camera at time $t+1$. In our case we have three cameras.
The dataset is composed of $N$ episodes with different lengths $l_i$:
\begin{align}\label{eq:dataset}
    \mathcal{D} = \{\{(a_t, I, p_{t+1} C_{\text{head}, t+1}, C_{\text{right\_wrist},t+1}, C_{\text{left\_wrist},t+1})\}_{t=1}^{l_i}\}_{i=1}^{N}.
\end{align}
We will consider imitation learning based on a behavior-cloning objective that maximizes the likelihood of an action given a state:
\begin{align}\label{eq:bc}
    \arg\max_\theta \mathbb{E}_{(a,s)\sim\mathcal{D}}[\log \pi_\theta(a|s)].
\end{align}

\subsection{Task Stages}
We introduce task stages, a subtask decomposition. Instead of relying only on a binary success criterion, which is often not very informative for low-performing policies, task stages give us fine-grained insight into which parts of a task are difficult for a policy. Furthermore, they also allow us to track task progress.
Our task stages can be seen as milestones. Each task starts at stage zero, and the stage can only increase, never decrease. Each stage has a set of constraints that must be satisfied, and the previous stage must be lower for it to become active.
Formally, each task has a set of stages $K=\{0, 1, \ldots, n\}$, where each stage has a set of constraints $C_k = \{c_{k,1}, \ldots, c_{k,m}\}$, with $c_{k,i}: S \rightarrow \{0, 1\}$ being a function that takes the current environment state and outputs whether the condition is satisfied. We further define a helper function $h(c, k): \{0, 1\}\times \mathbb{N}\rightarrow\mathbb{N}$, which returns $k$ if the condition $c$ is fulfilled and zero otherwise. Furthermore, the current stage is denoted by $k_t\in K$. The stage for the next timestep $t+1$ is then given by
\begin{align}
    \text{stage}_{t+1} = \max(\{\text{stage}_t\} \cup \{h( \wedge_{i} c_{k,i}(s_t), k): \forall k\in K\}).
\end{align}
For the case that both arms have independent sub-goals such that the stages could be modeled as a tree, which is e.g. the case in \binsort{}, we merge coexisting stages with \emph{exist} and \emph{all} quantifier conditions. For example one could specify that at least one arm needs to achieve a concrete goal in order for the agent to proceed to that stage.

The stages allow us to define the following two metrics which give meaningful values even when the task success rate is close to zero:
normalized mean progress per timestep $p_t=\frac{\mathbb{E}_e[\text{stage}_t^{(e)}] }{\max K}$ where $e$ is the episode; and 
normalized average final stage over all episodes $p_\text{final}=\frac{\mathbb{E}_e[\max_{t\in\mathcal{E}(e)}\text{stage}_t^{(e)}] }{\max K}$, where $\mathcal{E}$ maps to the number of time steps in $e$.
And finally, the probability that the policy fails in stage $k$ is given by $P(\max_{t\in\mathcal{E}(e)}\text{stage}_t^{(e)} = k)$.
The latter can be visualized as fractional bar plots that show in which stage the policy struggled the most, and the success rate can be expressed by the probability that the policy reached the final stage: $P(\max_{t\in\mathcal{E}(e)}\text{stage}_t^{(e)} = \max K)$.

We additionally report the \emph{stage-conditional pass rate}
\begin{align}
    \passk = P\bigl(\kappa \ge k+1 \mid \kappa \ge k\bigr), \qquad k=0,\ldots,n-1,
\end{align}
the probability that an episode which reached stage $k$ also reaches stage $k+1$. Unlike the failure-stage distribution, $\passk$ is normalized by the episodes that arrived at stage $k$, so a hard late stage is not masked by early attrition, and the success rate is the product of all pass rates, $P(\kappa=\max K)=\prod_k\passk$. We call $\arg\min_k\passk$ the \emph{bottleneck stage}. Success and pass rates are binomial proportions, and we report Wilson 95\% intervals~\cite{kressgazit2024empirical} for all of them.

\subsection{Franka Duo Setup}
\begin{figure}[t]
  \centering
  \includegraphics[width=0.45\linewidth]{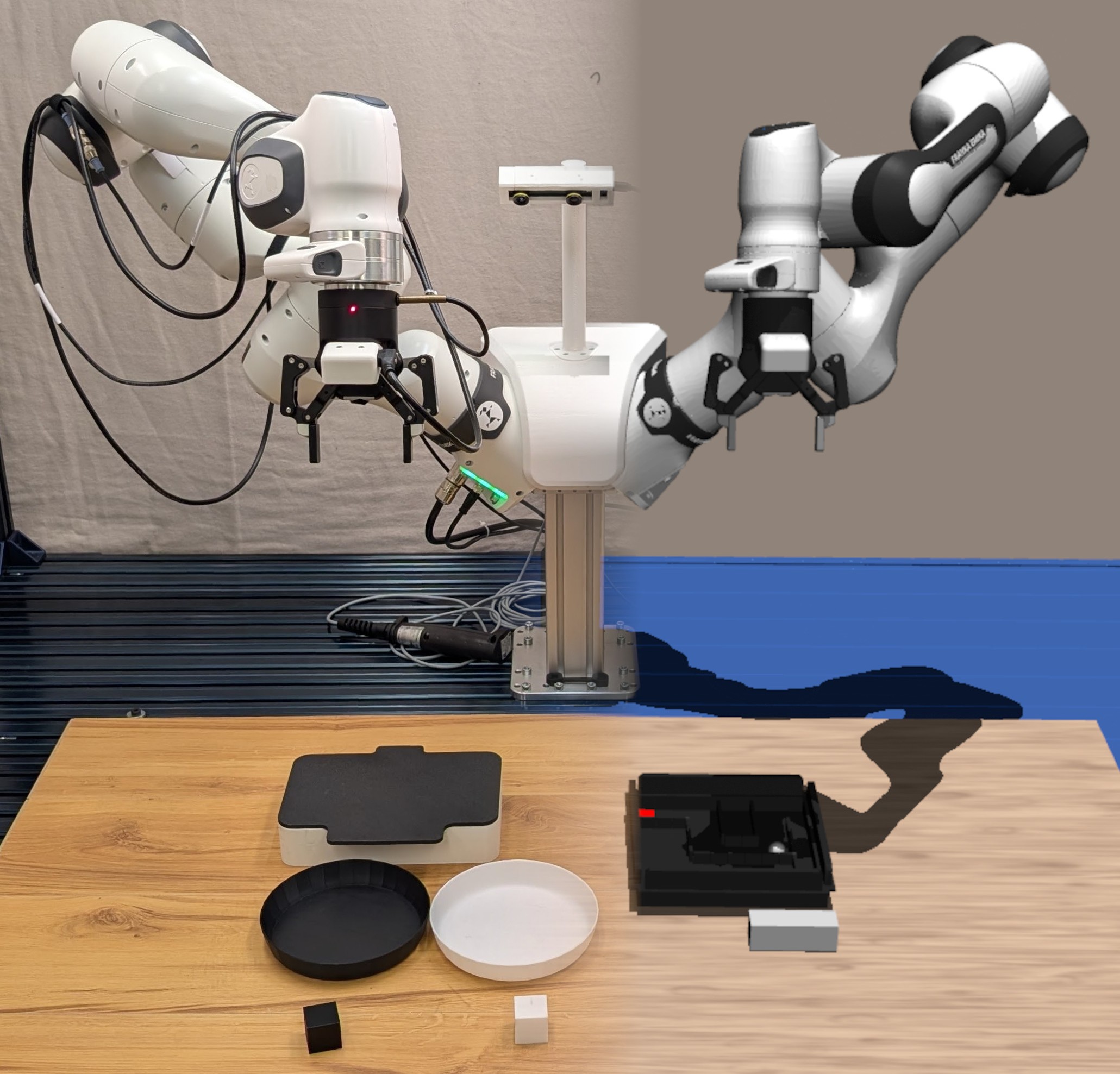}
  \caption{\frankaduo{} setup. Left shows the real-world setup with printed assets. Right shows the simulated MuJoCo scene.}%
  \label{fig:setup}
\end{figure}
\frankaduo{} is a novel dual-arm arrangement of \franka{} robotic arms defined by the manufacturer Franka Robotics.
The mounting configuration is chosen such that both robots' ISO cubes overlap, ensuring strong dual-arm manipulability.
Both arms have a two-finger Robotiq 2F-85 gripper attached.
The setup uses a main Zed Mini stereo camera with its left lens centered between the two arms and two RealSense D405 wrist cameras.
All mounting components, including the mounting block, its cover, the main camera holder, and the wrist camera mounts, can be purchased directly from the manufacturer.
For custom fabrication, STL files and full specifications
are provided in Franka's technical manuals~\cite{franka_docs}.

We reconstructed the same setup in a \mbox{MuJoCo}~\cite{mujoco} simulation with assets available from MuJoCo Menagerie~\cite{menagerie2022github} and the hardware manufacturer. \autoref{fig:setup} depicts the real-world setup and the simulated scene.

\begin{table}[ht]
\centering
\caption{Overview of all eleven tasks in our four bimanual task categories. ($n$) indicates the number of stages for each task. Tasks above the gray line are implemented in simulation and real-world settings. Tasks below are only implemented in simulation.}
\label{tab:taskcat}
\begin{tabular}{llll}
    \toprule
    \textbf{\asupp} & \textbf{\bimani} & \textbf{\sequhand} & \textbf{\paraexe} \\
    \midrule
    \hingechest\ (3)  & \ballmaze\ (4)     & \transfercube\ (4)     & \binsort\ (5) \\
    \arrayrulecolor{lightgray}
    \midrule
    \arrayrulecolor{black}
    \springdoor\ (3)  & \carrypot\ (3)     & \transferreorient\ (4) & \\
    \pourmarbles\ (6) & \blockbalance\ (5) & \transfergate\ (4)     & \\
                 & \joinblocks\ (3)   &                   & \\
    \bottomrule
\end{tabular}
\label{tab:bimanual_tasks}
\end{table}

\subsection{Bimanual Task Taxonomy}
We define a task taxonomy inspired by the hierarchical taxonomy of \citet{krebs} for bimanual tasks, but distinguishes tasks based on the functional roles the robot arms must fulfill to complete the task successfully.
Our taxonomy defines four non-hierarchical categories of bimanual manipulation tasks, organized by how the two arms divide functional roles. This structure allows us to evaluate which coordination patterns the tested policies handle well and where they struggle.
\\
\emph{\asupp}: One arm stabilizes or holds an object while the other arm interacts with it indirectly. For example, when placing a cube into a box with an attached lid, one arm must hold the lid open while the other inserts the cube. The task cannot be solved unimanually, as one arm is required to create the conditions necessary for success.
This category tests how well a policy understands and leverages its two arms in situations that require physical reasoning.\\
\emph{\bimani}: Both arms jointly manipulate the same object, forming a closed kinematic chain. A representative example is lifting a heavy pot using both arms. Neither arm can accomplish the task independently due to weight and stability constraints. Both arms must actively contribute rather than one serving a purely supportive role. These tasks evaluate how effectively the two arms coordinate under direct mutual physical interaction.\\
\emph{\sequhand}: An object is transferred between both arms due to workspace or task constraints. Cube transfer tasks fall into this category. Here the role of the acting arm changes at the handover from one arm to the other while both are interacting with each other. This category evaluates temporal coordination and the ability to manage handover phases.\\
\emph{\paraexe}: Both arms execute independent unimanual tasks, either simultaneously or sequentially. One or both arms may be active, but without direct interdependence. This setting evaluates whether dual-arm policies can maintain effective independent control, balance arm usage, and exhibit or avoid unintended dominance patterns.

In total, DuoBench comprises eleven tasks spanning the four categories. A summary of the tasks and their categories can be seen in \autoref{tab:bimanual_tasks}, and an image of each task is shown in \autoref{fig:overview}.
All tasks include at least one kind of randomization, e.g., random initial object position within a specified boundary; for details and the number of stages per task see \autoref{app:tasks}.
Most tasks are designed such that the required objects can be reproduced in real-world settings via 3D printing, while the unprintable objects are common household objects. Additionally, we replicated four tasks on our real-world \frankaduo{} lab setup, one from each category: \hingechest, \ballmaze, \transfercube, and \binsort. 

\subsection{Task Composition, Data Collection and Evaluation}
To facilitate task creation, the simulation benchmark exposes a generic task interface with three components: a task composer that uses MuJoCo's \texttt{MJSpec} interface to instantiate scene objects, a reset wrapper that handles randomized initialization, and a transparent stage wrapper that realizes the task-stage formalism introduced above. Both wrappers have direct access to the simulator state for implementing reset logic and evaluating task progress. Concretely, each stage is associated with a subtask instruction, an internal stage state, and a set of constraints. At every step, the wrapper evaluates the stage conditions, updates the current stage, and exposes success signals and stage-based rewards.
Each defined task is automatically registered with Gymnasium~\cite{gymnasium} via a unique ID, which then can create the complete wrapped \texttt{gym.Env} environment via the \texttt{gym.make} factory.

For real-world deployment, we use 3D-printed task assets, allowing us to reproduce the given simulation task exactly. Stage progress needs to be tracked manually in this case. The goal of reproducibility here is not to claim that policies trained on our real-world data directly generalize across setups, but rather to enable benchmark users to reproduce the tasks on their own hardware and collect comparable demonstrations for model comparison across labs.

Human-teleoperated data can be collected in both simulation and real-world settings using a virtual reality (VR) headset and the IRIS~\cite{iris} application. In simulation, the setup is projected in an augmented-reality fashion, while passthrough view is used in the real scenario. 
A data recording wrapper saves the observations and actions as defined in equation~\eqref{eq:dataset} during teleoperation. Additionally, in the simulation scenario, it also records the simulation state. Our episode replayer can use this data to repeat a given recording with new visual features, such as object colors and lighting conditions, allowing an existing simulation dataset to be ablated with ease.

Finally, we use VLAgents~\cite{vlagents} library for policy evaluation, which records all trials. The library directly uses our existing \texttt{gym.Env} environments to collect fine-grained progress data. It uses a seeded initialization strategy to ensure consistent conditions. 

\section{Results}
\label{sec:results}

We evaluate three representative policies that support dual-arm control on \duobench{}: \act{}~\cite{act} as the task-specific imitation-learning baseline and two recent generalist \vla{} models, \piofive{}~\cite{piofive} and \xvla{}~\cite{xvla}. We evaluate all policies with 30 action steps between model prompts, corresponding to one second of open-loop execution in our \SI{30}{\hertz} setup.
All policies output absolute joint targets for both arms and grippers, every episode starts from the same home configuration, and object positions are drawn from the demonstration distribution with a fixed seed in simulation and matched by hand in the real world.

\begin{table}[t]
\centering
\caption{DuoBench simulation evaluation, 300 rollouts per cell. Each cell reports the success rate (SR, \%) with its Wilson 95\% interval, the mean normalized task progress $p_\text{final}$, and the bottleneck stage with its pass rate $\passk$.}
\label{tab:sim}
\resizebox{\columnwidth}{!}{%
\begin{tabular}{lrclrclrcl}
\toprule
Task & \multicolumn{3}{c}{\act} & \multicolumn{3}{c}{\piofive} & \multicolumn{3}{c}{\xvla} \\
\cmidrule(lr){2-4}\cmidrule(lr){5-7}\cmidrule(lr){8-10}
 & SR [CI] & $p_\text{final}$ & bottleneck stage ($\rho_k$) & SR [CI] & $p_\text{final}$ & bottleneck stage ($\rho_k$) & SR [CI] & $p_\text{final}$ & bottleneck stage ($\rho_k$) \\
\midrule
\multicolumn{10}{l}{\textit{\asupp}} \\
\hingechest & 26.7\,\textcolor{gray}{\scriptsize[22,32]} & \textcolor{gray}{0.55} & cube + lid \textcolor{gray}{\scriptsize(0.39)} & 26.7\,\textcolor{gray}{\scriptsize[22,32]} & \textcolor{gray}{0.39} & lid or cube \textcolor{gray}{\scriptsize(0.56)} & 12.0\,\textcolor{gray}{\scriptsize[9,16]} & \textcolor{gray}{0.30} & cube + lid \textcolor{gray}{\scriptsize(0.37)} \\
\springdoor & 22.3\,\textcolor{gray}{\scriptsize[18,27]} & \textcolor{gray}{0.50} & box + door \textcolor{gray}{\scriptsize(0.33)} & 19.7\,\textcolor{gray}{\scriptsize[16,25]} & \textcolor{gray}{0.44} & box + door \textcolor{gray}{\scriptsize(0.39)} & 13.7\,\textcolor{gray}{\scriptsize[10,18]} & \textcolor{gray}{0.38} & box + door \textcolor{gray}{\scriptsize(0.37)} \\
\pourmarbles & 0.3\,\textcolor{gray}{\scriptsize[0,2]} & \textcolor{gray}{0.12} & pour all \textcolor{gray}{\scriptsize(0.23)} & 0.0\,\textcolor{gray}{\scriptsize[0,1]} & \textcolor{gray}{0.08} & pour all \textcolor{gray}{\scriptsize(0.00)} & 0.0\,\textcolor{gray}{\scriptsize[0,1]} & \textcolor{gray}{0.07} & place cups \textcolor{gray}{\scriptsize(0.00)} \\
\multicolumn{10}{l}{\textit{\bimani}} \\
\carrypot & 54.0\,\textcolor{gray}{\scriptsize[48,60]} & \textcolor{gray}{0.70} & grasp + lift \textcolor{gray}{\scriptsize(0.55)} & 26.0\,\textcolor{gray}{\scriptsize[21,31]} & \textcolor{gray}{0.51} & grasp + lift \textcolor{gray}{\scriptsize(0.30)} & 12.3\,\textcolor{gray}{\scriptsize[9,17]} & \textcolor{gray}{0.40} & grasp + lift \textcolor{gray}{\scriptsize(0.14)} \\
\ballmaze & 8.0\,\textcolor{gray}{\scriptsize[5,12]} & \textcolor{gray}{0.71} & ball to goal \textcolor{gray}{\scriptsize(0.09)} & 5.0\,\textcolor{gray}{\scriptsize[3,8]} & \textcolor{gray}{0.67} & ball to goal \textcolor{gray}{\scriptsize(0.06)} & 2.0\,\textcolor{gray}{\scriptsize[1,4]} & \textcolor{gray}{0.71} & ball to goal \textcolor{gray}{\scriptsize(0.02)} \\
\blockbalance & 0.0\,\textcolor{gray}{\scriptsize[0,1]} & \textcolor{gray}{0.12} & grasp rects \textcolor{gray}{\scriptsize(0.00)} & 0.0\,\textcolor{gray}{\scriptsize[0,1]} & \textcolor{gray}{0.06} & grasp rects \textcolor{gray}{\scriptsize(0.00)} & 0.0\,\textcolor{gray}{\scriptsize[0,1]} & \textcolor{gray}{0.15} & release \textcolor{gray}{\scriptsize(0.00)} \\
\joinblocks & 2.3\,\textcolor{gray}{\scriptsize[1,5]} & \textcolor{gray}{0.17} & connect \textcolor{gray}{\scriptsize(0.06)} & 0.0\,\textcolor{gray}{\scriptsize[0,1]} & \textcolor{gray}{0.16} & to wall \textcolor{gray}{\scriptsize(0.00)} & 0.0\,\textcolor{gray}{\scriptsize[0,1]} & \textcolor{gray}{0.20} & to wall \textcolor{gray}{\scriptsize(0.00)} \\
\multicolumn{10}{l}{\textit{\sequhand}} \\
\transfercube & 10.0\,\textcolor{gray}{\scriptsize[7,14]} & \textcolor{gray}{0.14} & pick cube \textcolor{gray}{\scriptsize(0.20)} & 13.0\,\textcolor{gray}{\scriptsize[10,17]} & \textcolor{gray}{0.28} & pick cube \textcolor{gray}{\scriptsize(0.47)} & 9.3\,\textcolor{gray}{\scriptsize[7,13]} & \textcolor{gray}{0.17} & pick cube \textcolor{gray}{\scriptsize(0.33)} \\
\transfergate & 42.7\,\textcolor{gray}{\scriptsize[37,48]} & \textcolor{gray}{0.49} & pick box \textcolor{gray}{\scriptsize(0.59)} & 23.3\,\textcolor{gray}{\scriptsize[19,28]} & \textcolor{gray}{0.35} & pick box \textcolor{gray}{\scriptsize(0.49)} & 26.3\,\textcolor{gray}{\scriptsize[22,32]} & \textcolor{gray}{0.36} & pick box \textcolor{gray}{\scriptsize(0.51)} \\
\transferreorient & 0.3\,\textcolor{gray}{\scriptsize[0,2]} & \textcolor{gray}{0.35} & insert \textcolor{gray}{\scriptsize(0.01)} & 1.3\,\textcolor{gray}{\scriptsize[1,3]} & \textcolor{gray}{0.36} & insert \textcolor{gray}{\scriptsize(0.04)} & 0.0\,\textcolor{gray}{\scriptsize[0,1]} & \textcolor{gray}{0.30} & insert \textcolor{gray}{\scriptsize(0.00)} \\
\multicolumn{10}{l}{\textit{\paraexe}} \\
\binsort & 0.7\,\textcolor{gray}{\scriptsize[0,2]} & \textcolor{gray}{0.11} & pick 2nd \textcolor{gray}{\scriptsize(0.09)} & 1.7\,\textcolor{gray}{\scriptsize[1,4]} & \textcolor{gray}{0.14} & pick 2nd \textcolor{gray}{\scriptsize(0.16)} & 1.0\,\textcolor{gray}{\scriptsize[0,3]} & \textcolor{gray}{0.13} & pick 2nd \textcolor{gray}{\scriptsize(0.21)} \\
\bottomrule
\end{tabular}

}
\end{table}

\subsection{Datasets}
For each task, we provide 50 human-teleoperated demonstrations, comprising eleven simulation tasks and four real-world tasks for a total of 750 episodes, 442{,}907 frames, and an average episode length of about 350 frames or 12 seconds. Data frames are recorded at \SI{30}{\hertz}. Each observation contains a language instruction provided by the teleoperator, images from three cameras at a native resolution of $1280\times720$, joint states, Cartesian poses, and the full \franka{} robot states in the real world and the partial MuJoCo object state in simulation for replaying, while actions are collected as Cartesian commands through VR teleoperation. The raw recordings are stored in the \rcs{} parquet-based format and additionally converted to the LeRobot format for downstream training. During conversion, images are resized to $224\times224$, states and actions are flattened, and actions are mapped to joint space using observation-initialized inverse kinematics to avoid configuration ambiguities.

\subsection{Evaluation in Simulation}
\begin{figure}[t]
	\centering
	\includegraphics[width=\linewidth]{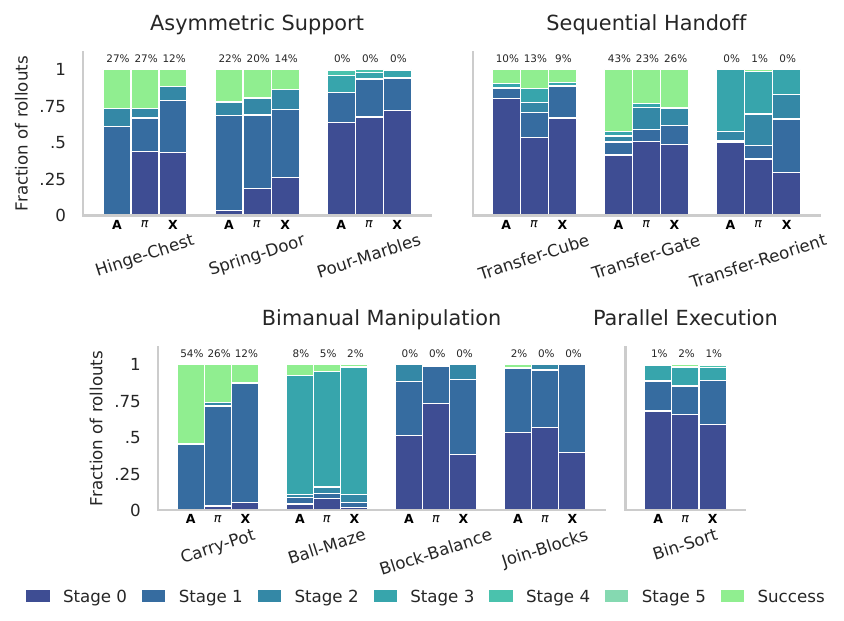}
	\caption{Fraction of rollouts in simulation that failed in a given stage. Success rates are annotated above. We abbreviate policy names as follows: \act{} as ``A'', \piofive{} as ``$\pi$'' and \xvla{} as ``X''.}
	\label{fig:stages:sim}
\end{figure}

We trained all three policies on the teleoperated data from all eleven simulation tasks. \act{} was trained on 50 episodes from each task individually, whereas the VLAs were trained on all 550 episodes with task instruction conditioning. We evaluate them in the same Gymnasium environments used for data collection. We did not use replayer-based data augmentation for comparability with real-world evaluation. Object positions are sampled from a consistent seed to provide equal conditions.
Each model is evaluated for 300 rollouts per task. Maximum cut-off lengths are calibrated per task according to the dataset statistics.
The stage is returned by the Gymnasium environment at each step in the form of reward and success, and allows us to compare policy performance even when the success rate is close to zero.
Both the success rates and $p_\text{final}$ are shown in \autoref{tab:sim}.
\autoref{fig:stages:sim} shows the fraction of runs that failed in a given stage. The largest fraction represents the stage where most runs failed and, thus, gives an indication of the part the policy struggles with.

Across policies, the stage distributions reveal that many failures already occur in the earliest stages, indicating that initial grasp acquisition and task setup remain dominant bottlenecks even before more complex coordination is required. This pattern is particularly informative in the low-data benchmark regime of only 50 demonstrations per task, where differences in data efficiency become visible. At the same time, \duobench{} distinguishes between tasks that appear similar at a high level but differ substantially in execution difficulty. For example, \transfergate{} is easier than \transfercube{} despite the additional scene structure, indicating that spatially constraining the handover can simplify coordination by making the transfer configuration more predictable. A notable result is the comparatively low performance on \binsort{}, which suggests that current policies struggle with parallel execution of both arms and with learning from demonstrations that admit multiple valid solution paths, such as different grasp or execution orders that do not affect task success. Finally, \act{} achieves the strongest performance on several tasks, including \transfergate{} and \carrypot{}. Beyond success rates, the stage-based analysis shows where each policy fails: in \transferreorient{}, for example, \xvla{} picks up the peg more often than \act{} but rarely passes it to the second gripper.
In terms of pass rates (\autoref{tab:sim}), the bottleneck is a stage that requires establishing a grasp in 17 of the 33 task-policy combinations. In all six combinations of \carrypot{} and \ballmaze{}, it is instead the motion that follows, where both arms move the same object together and thereby close a kinematic chain through it. The full pass-rate profiles of all tasks are given in \autoref{fig:passrates:sim} in the appendix.

\subsection{Evaluation in Real-World}
We also trained the policies on the data collected on the four replicated real-world tasks. Again, \act{} was trained per task, whereas the VLAs were trained on all 200 episodes combined. We evaluated each policy on each task for 15 rollouts. Intermediate stages and success are judged by a human operator.
The resulting success rates and $p_\text{final}$ can be seen in \autoref{tab:mix} under the datasets ``Real''.

The real-world results are largely consistent with the simulation findings in that failures are still dominated by the earliest interaction phases, especially grasp acquisition. At the same time, the real-world experiments suggest that, once grasping succeeds, later stages are often completed reliably: for example, in \transfercube{} the handover itself is frequently successful once the object is securely grasped, and in \binsort{} the placement stage is typically not the limiting factor. In \ballmaze{}, the policies further show that they can reproduce the required dual-arm contact pattern, but often fail to act on the underlying task physics; for instance, a policy may grasp the maze correctly without lifting and tilting it in a way that would move the ball toward the goal. Overall, the real-world evaluation supports the same broad conclusions as the simulation benchmark while highlighting that physically grounded interaction remains a central challenge even in scenes closely matching the training setup.

\begin{figure}[t]
	\centering
	\includegraphics[width=0.85\linewidth]{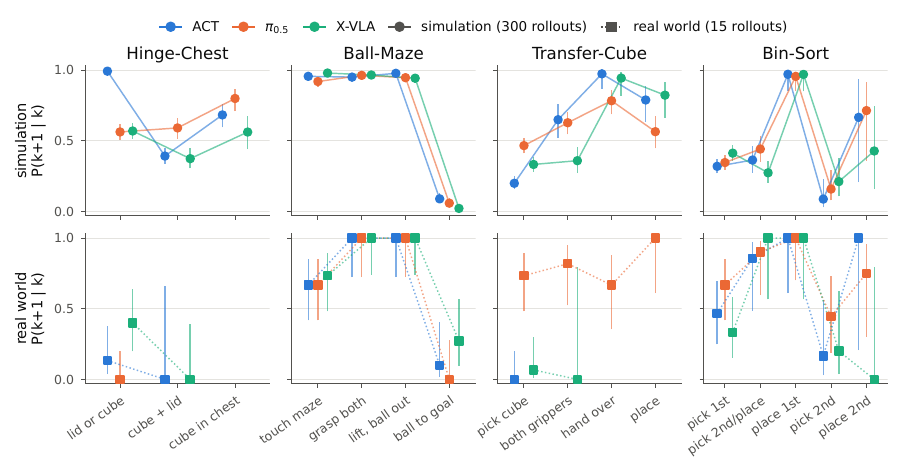}
	\vspace{-0.5\baselineskip}
	\caption{Stage-conditional pass rates $\passk$ with Wilson 95\% intervals for the four tasks that exist in simulation (top, 300 rollouts) and the real world (bottom, 15 rollouts), each policy trained on the data of the evaluated domain. The $x$-axis names the sub-instruction to leave stage $k$. A series ends where no episode reached the next stage.}
	\label{fig:passrates}
	\vspace{-0.8\baselineskip}
\end{figure}

\autoref{fig:passrates} compares the pass-rate profiles of the four replicated tasks across domains. In \ballmaze{}, the bimanual grasp is reliable in both domains, and guiding the ball to the goal is the bottleneck. In \transfercube{}, picking up the cube is the bottleneck in simulation for all policies and on the real robot for \act{} and \xvla{}, whereas \piofive{} picks more reliably on the real robot than in simulation. Simulation thus locates bottlenecks faithfully even where it misestimates success rates.

\subsection{Mixed Training and Sim-to-Real Gap}
Finally, we also trained the policies on the complete simulation-and-real-data mix to test the sim-to-real gap. \act{} was trained per task on simulation and real-world data for the four real-world tasks, while the VLAs were trained on the full 750-episode dataset.
The resulting success rates and $p_\text{final}$ can be seen in \autoref{tab:mix} under the datasets ``Mixed''. For comparison, the table also shows the success rates of the model when trained only on the evaluated domain.

The mixed-training results show a clear remaining sim-to-real gap across all evaluated policies, confirming that alignment between the simulated and physical benchmark settings does not by itself eliminate cross-domain transfer challenges. At the same time, joint training on simulation and real-world data can improve real-world performance for some models without requiring separate domain-specific training. This effect is most visible for \act{} and \piofive{}, which improve or maintain their real-world performance, although the gains are not uniform across all tasks and models.
In simulation, mixed training largely preserves the performance of \act{}, but degrades \xvla{} and \piofive{}, the latter mainly on \hingechest{}, indicating that naively combining both domains is not universally beneficial; for \xvla{}, one possible reason is that we did not use distinct domain IDs for cross-domain finetuning. Overall, these results suggest that DuoBench is suitable not only for evaluating policies within each domain, but also for studying how different training strategies trade off simulation performance, real-world robustness, and cross-domain transfer.

\begin{table}[t]
\centering
\caption{Evaluation of domain independent training vs mixed-data training. Each cell reports success rate with mean normalized task progress in parentheses for real and simulation evaluation.}
\label{tab:mix}
\resizebox{\columnwidth}{!}{%
\begin{tabular}{lcccccccccccc}
\toprule
Model & \multicolumn{4}{c}{\act} & \multicolumn{4}{c}{\piofive} & \multicolumn{4}{c}{\xvla} \\
\cmidrule(lr){2-5}\cmidrule(lr){6-9}\cmidrule(lr){10-13}
Train dataset & Sim & Real & \multicolumn{2}{c}{Mixed} & Sim & Real & \multicolumn{2}{c}{Mixed} & Sim & Real & \multicolumn{2}{c}{Mixed} \\
\cmidrule(lr){2-2}\cmidrule(lr){3-3}\cmidrule(lr){4-5}\cmidrule(lr){6-6}\cmidrule(lr){7-7}\cmidrule(lr){8-9}\cmidrule(lr){10-10}\cmidrule(lr){11-11}\cmidrule(lr){12-13}
Eval domain & Sim & Real & Sim & Real & Sim & Real & Sim & Real & Sim & Real & Sim & Real \\
\midrule
\hingechest & 26.7 \textcolor{gray}{(0.55)} & 0 \textcolor{gray}{(0.04)} & 26.7 \textcolor{gray}{(0.56)} & 0 \textcolor{gray}{(0.18)} & 26.7 \textcolor{gray}{(0.39)} & 0 \textcolor{gray}{(0.00)} & 4.7 \textcolor{gray}{(0.08)} & 0 \textcolor{gray}{(0.07)} & 12.0 \textcolor{gray}{(0.30)} & 0 \textcolor{gray}{(0.13)} & 8.7 \textcolor{gray}{(0.19)} & 0 \textcolor{gray}{(0.04)} \\
\ballmaze & 8.0 \textcolor{gray}{(0.71)} & 7 \textcolor{gray}{(0.52)} & 3.7 \textcolor{gray}{(0.69)} & 13 \textcolor{gray}{(0.43)} & 5.0 \textcolor{gray}{(0.67)} & 0 \textcolor{gray}{(0.50)} & 6.0 \textcolor{gray}{(0.69)} & 0 \textcolor{gray}{(0.30)} & 2.0 \textcolor{gray}{(0.71)} & 20 \textcolor{gray}{(0.60)} & 1.3 \textcolor{gray}{(0.66)} & 7 \textcolor{gray}{(0.12)} \\
\transfercube & 10.0 \textcolor{gray}{(0.14)} & 0 \textcolor{gray}{(0.00)} & 8.3 \textcolor{gray}{(0.13)} & 0 \textcolor{gray}{(0.00)} & 13.0 \textcolor{gray}{(0.28)} & 40 \textcolor{gray}{(0.53)} & 8.0 \textcolor{gray}{(0.23)} & 53 \textcolor{gray}{(0.60)} & 9.3 \textcolor{gray}{(0.17)} & 0 \textcolor{gray}{(0.02)} & 4.0 \textcolor{gray}{(0.15)} & 7 \textcolor{gray}{(0.13)} \\
\binsort & 0.7 \textcolor{gray}{(0.11)} & 7 \textcolor{gray}{(0.28)} & 0.7 \textcolor{gray}{(0.15)} & 13 \textcolor{gray}{(0.37)} & 1.7 \textcolor{gray}{(0.14)} & 20 \textcolor{gray}{(0.47)} & 0.7 \textcolor{gray}{(0.14)} & 7 \textcolor{gray}{(0.29)} & 1.0 \textcolor{gray}{(0.13)} & 0 \textcolor{gray}{(0.21)} & 0.3 \textcolor{gray}{(0.10)} & 7 \textcolor{gray}{(0.20)} \\
\midrule
\textbf{Overall} & 11.3 \textcolor{gray}{(0.38)} & 3 \textcolor{gray}{(0.21)} & 9.8 \textcolor{gray}{(0.38)} & 7 \textcolor{gray}{(0.25)} & 11.6 \textcolor{gray}{(0.37)} & 15 \textcolor{gray}{(0.37)} & 4.8 \textcolor{gray}{(0.28)} & 15 \textcolor{gray}{(0.31)} & 6.1 \textcolor{gray}{(0.33)} & 5 \textcolor{gray}{(0.24)} & 3.6 \textcolor{gray}{(0.28)} & 5 \textcolor{gray}{(0.12)} \\
\bottomrule
\end{tabular}

}
\end{table}

\clearpage
\section{Limitations}
\label{sec:limitations}
    While DuoBench provides a challenging evaluation framework, there are aspects that warrant future effort. First, real-world evaluation provides weaker diagnostic signals than simulation, as tracking stage progress remains a manual process in this setting. Second, the task taxonomy and stage definitions are hand-crafted and therefore reflect design choices about which coordination patterns and failure modes are emphasized. Finally, the current benchmark assumes a fixed hardware and sensing setup, leaving other configurations to future work.

\section{Conclusion}
\label{sec:conclusion}

In this work, we introduced DuoBench, an extensible benchmarking framework for bimanual manipulation with eleven tasks across four coordination categories in simulation and partially in the real world. Alongside reproducible task definitions and human-teleoperated datasets, DuoBench contributes a coordination taxonomy and stage-based evaluation that enable fine-grained analysis of semantic failure modes beyond binary success.
Our experiments show that bimanual manipulation remains challenging for current policies, with recurring difficulties in early interaction stages, parallel arm execution, and cross-domain transfer between simulation and real-world settings.
In the future, we plan to extend DuoBench by adding more task varieties, particularly in tactile-relevant domains. We hope that DuoBench will provide a useful foundation for future work on dual-arm policy learning, richer sensing modalities, and more robust cross-domain training and evaluation.

\acknowledgments{We would like to thank Devadas Vijayan Sheela for helping us with the real-world evaluation. This work has been partially supported by the project GeniusRobot and funded by the German Federal Ministry of Education and Research (BMBF grant no.~01IS24083).
It has also been partially supported by the German Federal Ministry of Research, Technology and Space (BMFTR) under the Robotics Institute Germany (RIG).
The authors acknowledge the HPC resources provided by the Erlangen National HPC Center (NHR@FAU) under the BayernKI project no.~v106be.
}

\bibliography{bibliography}

@string{neurips = "Advances in Neural Information Processing Systems"}

@string{ieeeral = "IEEE Robotics and Automation Letters"}

@string{ieeepami = "IEEE Transactions on Pattern Analysis and Machine Intelligence"}

@string{IROS = "Proc.~of the IEEE/RSJ Int.~Conf.~on Intelligent Robots and Systems (IROS)"}

@string{ICRA = "Proc.~of the IEEE Int.~Conf.~on Robotics \& Automation (ICRA)"}

@string{ICCV = "Proc.~of Int.~Conf.~on Computer Vision (ICCV)"}

@string{ICML = "Proc.~of the Int.~Conf.~on Machine Learning (ICML)"}

@string{CVPR = "Proc.~of the IEEE Computer Society Conference on
                  Computer Vision and Pattern Recognition (CVPR)"}

@string{IJRR = "Int.~Journal of Robotics Research (IJRR)"}

@string{rss = "Proc.~of Robotics: Science and Systems (RSS)"}

@string{ICLR = "Proc.~of the Int.~Conf.~on Learning Representations (ICLR)"}

@string{CORL = "Proc.~of the Conf.~on Robot Learning (CoRL)"}

@misc{lerobot,
    author = {Cadene, Remi and Alibert, Simon and Soare, Alexander and Gallouedec, Quentin and Zouitine, Adil and Palma, Steven and Kooijmans, Pepijn and Aractingi, Michel and Shukor, Mustafa and Aubakirova, Dana and Russi, Martino and Capuano, Francesco and Pascal, Caroline and Choghari, Jade and Moss, Jess and Wolf, Thomas},
    title = {{LeRobot}: State-of-the-art Machine Learning for Real-World Robotics in Pytorch},
    howpublished = {\url{https://github.com/huggingface/lerobot}},
    year = {2024}
}

@software{isaacsim,
    author = {{NVIDIA}},
    license = {Apache-2.0},
    title = {{Isaac Sim}},
    url = {https://github.com/isaac-sim/IsaacSim},
    version = {5.0.0}
}

@inproceedings{mujoco,
  author={Todorov, Emanuel and Erez, Tom and Tassa, Yuval},
  booktitle= IROS, 
  title={MuJoCo: A physics engine for model-based control}, 
  year={2012}
}

@misc{robosuite,
      title={robosuite: A Modular Simulation Framework and Benchmark for Robot Learning}, 
      author={Yuke Zhu and Josiah Wong and Ajay Mandlekar and Roberto Martín-Martín and Abhishek Joshi and Kevin Lin and Abhiram Maddukuri and Soroush Nasiriany and Yifeng Zhu},
      year={2025},
      howpublished={\url{https://arxiv.org/abs/2009.12293}}
}

@inproceedings{libero,
  title={Libero: Benchmarking knowledge transfer for lifelong robot learning},
  author={Liu, Bo and Zhu, Yifeng and Gao, Chongkai and Feng, Yihao and Liu, Qiang and Zhu, Yuke and Stone, Peter},
  booktitle=neurips,
  year={2023}
}

@inproceedings{metaworld,
  title = 	 {{Meta-World}: A Benchmark and Evaluation for Multi-Task and Meta Reinforcement Learning},
  author =       {Yu, Tianhe and Quillen, Deirdre and He, Zhanpeng and Julian, Ryan and Hausman, Karol and Finn, Chelsea and Levine, Sergey},
  booktitle = CORL,
  year = 	 {2020}
}

@inproceedings{sapien,
    author = {Xiang, Fanbo and Qin, Yuzhe and Mo, Kaichun and Xia, Yikuan and Zhu, Hao and Liu, Fangchen and Liu, Minghua and Jiang, Hanxiao and Yuan, Yifu and Wang, He and Yi, Li and Chang, Angel X. and Guibas, Leonidas J. and Su, Hao},
    title = {{SAPIEN}: A SimulAted Part-Based Interactive ENvironment},
    booktitle = CVPR,
    year = {2020}
}

@inproceedings{maniskill,
  title = {{{ManiSkill}}: {{Generalizable Manipulation Skill Benchmark}} with {{Large-Scale Demonstrations}}},
  author = {Mu, Tongzhou and Ling, Zhan and Xiang, Fanbo and Yang, Derek Cathera and Li, Xuanlin and Tao, Stone and Huang, Zhiao and Jia, Zhiwei and Su, Hao},
  year = {2021},
  booktitle = {{{Neural Information Processing Systems Datasets}} and {{Benchmarks Track}} ({{Round}} 2)},
}

@inproceedings{maniskill2,
    title={{ManiSkill2}: A Unified Benchmark for Generalizable Manipulation Skills},
    author={Jiayuan Gu and Fanbo Xiang and Xuanlin Li and Zhan Ling and Xiqiang Liu and Tongzhou Mu and Yihe Tang and Stone Tao and Xinyue Wei and Yunchao Yao and Xiaodi Yuan and Pengwei Xie and Zhiao Huang and Rui Chen and Hao Su},
    booktitle= ICLR,
    year={2023}
}

@inproceedings{maniskill3,
    author = {Stone, Tao and Xiang, Fanbo and Shukla, Arth and Qin, Yuzhe and Hinrichsen, Xander and Yuan, Xiaodi and Bao, Chen and Lin, Xinsong and Liu, Yulin and Chan, Tse-Kai and Gao, Yuan and Li, Xuanlin and Mu, Tongzhou and Xiao, Nan and Gurha, Arnav and N., Viswesh and Choi, Yong Woo and Chen, Yen-Ru and Huang, Zhiao and Calandra, Roberto and Chen, Rui and Luo, Shan and Su, Hao},
    title = {Demonstrating GPU Parallelized Robot Simulation and Rendering for Generalizable Embodied AI with {ManiSkill3}},
    booktitle = rss,
    year = 2025
}

@article{calvin,
  author={Mees, Oier and Hermann, Lukas and Rosete-Beas, Erick and Burgard, Wolfram},
  journal=ieeeral, 
  title={{CALVIN}: A Benchmark for Language-Conditioned Policy Learning for Long-Horizon Robot Manipulation Tasks}, 
  year={2022},
  volume={7},
  number={3}
}

@inproceedings{ikea,
  title = {{{IKEA Furniture Assembly Environment}} for {{Long-Horizon Complex Manipulation Tasks}}},
  booktitle = ICRA,
  author = {Lee, Youngwoon and Hu, Edward S. and Lim, Joseph J.},
  year = 2021
}

@inproceedings{behavior,
  title = {{{BEHAVIOR}}: {{Benchmark}} for {{Everyday Household Activities}} in {{Virtual}}, {{Interactive}}, and {{Ecological Environments}}},
  booktitle = CORL,
  author = {Srivastava, Sanjana and Li, Chengshu and Lingelbach, Michael and {Mart{\'i}n-Mart{\'i}n}, Roberto and Xia, Fei and Vainio, Kent Elliott and Lian, Zheng and Gokmen, Cem and Buch, Shyamal and Liu, Karen and Savarese, Silvio and Gweon, Hyowon and Wu, Jiajun and {Fei-Fei}, Li},
  year = {2022}
}

@inproceedings{vima,
  title = {{{VIMA}}: {{Robot Manipulation}} with {{Multimodal Prompts}}},
  booktitle = ICML,
  author = {Jiang, Yunfan and Gupta, Agrim and Zhang, Zichen and Wang, Guanzhi and Dou, Yongqiang and Chen, Yanjun and {Fei-Fei}, Li and Anandkumar, Anima and Zhu, Yuke and Fan, Linxi},
  year = 2023
}

@inproceedings{robohive,
  title = {{{RoboHive}}: {{A Unified Framework}} for {{Robot Learning}}},
  author = {Kumar, Vikash and Shah, Rutav and Zhou, Gaoyue and Moens, Vincent and Caggiano, Vittorio and Gupta, Abhishek and Rajeswaran, Aravind},
  year = 2023,
  booktitle = neurips
}

@article{bidexhands,
  author={Chen, Yuanpei and Geng, Yiran and Zhong, Fangwei and Ji, Jiaming and Jiang, Jiechuang and Lu, Zongqing and Dong, Hao and Yang, Yaodong},
  journal=ieeepami, 
  title={{Bi-DexHands}: Towards Human-Level Bimanual Dexterous Manipulation}, 
  year={2024},
  volume={46},
  number={5}
}

@inproceedings{humanoidbench,
  title = {{{HumanoidBench}}: {{Simulated}} Humanoid Benchmark for Whole-Body Locomotion and Manipulation},
  booktitle = rss,
  author = {Sferrazza, Carmelo and Huang, Dun-Ming and Lin, Xingyu and Lee, Youngwoon and Abbeel, Pieter},
  year = 2024
}

@inproceedings{roboarena,
  title = {{{RoboArena}}: {{Distributed Real-World Evaluation}} of {{Generalist Robot Policies}}},
  booktitle = CORL,
  author = {Atreya, Pranav and Pertsch, Karl and Lee, Tony and Kim, Moo Jin and Jain, Arhan and Kuramshin, Artur and Neary, Cyrus and Hu, Edward S. and Arora, Kanav and Ellis, Kirsty and Macesanu, Luca and Leonard, Matthew and Cho, Meedeum and Aslan, Ozgur and Dass, Shivin and Wang, Tony and Yuan, Xingfang and Gupta, Abhishek and Jayaraman, Dinesh and Berseth, Glen and Daniilidis, Kostas and {Mart{\'i}n-Mart{\'i}n}, Roberto and Lee, Youngwoon and Liang, Percy and Finn, Chelsea and Levine, Sergey},
  year = 2025
}

@inproceedings{bigym,
  title = {{{BiGym}}: {{A Demo-Driven Mobile Bi-Manual Manipulation Benchmark}}},
  booktitle = CORL,
  author = {Chernyadev, Nikita and Backshall, Nicholas and Ma, Xiao and Lu, Yunfan and Seo, Younggyo and James, Stephen},
  year = 2025
}

@misc{liberoplus,
  title = {{{LIBERO-Plus}}: {{In-depth Robustness Analysis}} of {{Vision-Language-Action Models}}},
  author = {Fei, Senyu and Wang, Siyin and Shi, Junhao and Dai, Zihao and Cai, Jikun and Qian, Pengfang and Ji, Li and He, Xinzhe and Zhang, Shiduo and Fei, Zhaoye and Fu, Jinlan and Gong, Jingjing and Qiu, Xipeng},
  year = 2025,
  howpublished = {\url{https://arxiv.org/abs/2510.13626}}
}

@misc{liberox,
  title = {{{LIBERO-X}}: {{Robustness Litmus}} for {{Vision-Language-Action Models}}},
  author = {Wang, Guodong and Zhang, Chenkai and Liu, Qingjie and Zhang, Jinjin and Cai, Jiancheng and Liu, Junjie and Liu, Xinmin},
  year = 2026,
  howpublished = {\url{https://arxiv.org/abs/2602.06556}}
}

@misc{liberopro,
  title = {{{LIBERO-PRO}}: {{Towards Robust}} and {{Fair Evaluation}} of {{Vision-Language-Action Models Beyond Memorization}}},
  author = {Zhou, Xueyang and Xu, Yangming and Tie, Guiyao and Chen, Yongchao and Zhang, Guowen and Chu, Duanfeng and Zhou, Pan and Sun, Lichao},
  year = 2025,
  howpublished = {\url{https://arxiv.org/abs/2510.03827}}
}

@inproceedings{robocasa,
  title={{RoboCasa}: Large-Scale Simulation of Household Tasks for Generalist Robots},
  author={Soroush Nasiriany and Abhiram Maddukuri and Lance Zhang and Adeet Parikh and Aaron Lo and Abhishek Joshi and Ajay Mandlekar and Yuke Zhu},
  booktitle=rss,
  year={2024}
}

@inproceedings{robocasa365,
  title={{RoboCasa365}: A Large-Scale Simulation Framework for Training and Benchmarking Generalist Robots}, 
  author={Soroush Nasiriany and Sepehr Nasiriany and Abhiram Maddukuri and Yuke Zhu},
  booktitle=ICLR,
  year={2026}
}

@article{rlbench,
  author={James, Stephen and Ma, Zicong and Arrojo, David Rovick and Davison, Andrew J.},
  journal= ieeeral, 
  title={{RLBench}: The Robot Learning Benchmark \& Learning Environment}, 
  year={2020},
  volume={5},
  number={2}
}

@inproceedings{gymnasium,
      title={Gymnasium: A Standard Interface for Reinforcement Learning Environments}, 
      author={Mark Towers and Ariel Kwiatkowski and Jordan Terry and John U. Balis and Gianluca De Cola and Tristan Deleu and Manuel Goulão and Andreas Kallinteris and Markus Krimmel and Arjun KG and Rodrigo Perez-Vicente and Andrea Pierré and Sander Schulhoff and Jun Jet Tai and Hannah Tan and Omar G. Younis},
      booktitle=neurips,
      year={2025}
}

@INPROCEEDINGS{droiddataset, 
    AUTHOR    = {Alexander Khazatsky AND Karl Pertsch AND Suraj Nair AND Ashwin Balakrishna AND Sudeep Dasari AND Siddharth Karamcheti AND Soroush Nasiriany AND Mohan Kumar Srirama AND Lawrence Yunliang Chen AND Kirsty Ellis AND Peter David Fagan AND Joey Hejna AND Masha Itkina AND Marion Lepert AND Yecheng Jason Ma AND Patrick Tree Miller AND Jimmy Wu AND Suneel Belkhale AND Shivin Dass AND Huy Ha AND Arhan Jain AND Abraham Lee AND Youngwoon Lee AND Marius Memmel AND Sungjae Park AND Ilija Radosavovic AND Kaiyuan Wang AND Albert Zhan AND Kevin Black AND Cheng Chi AND Kyle Beltran Hatch AND Shan Lin AND Jingpei Lu AND Jean Mercat AND Abdul Rehman AND Pannag R Sanketi AND Archit Sharma AND Cody Simpson AND Quan Vuong AND Homer Rich Walke AND Blake Wulfe AND Ted Xiao AND Jonathan Heewon Yang AND Arefeh Yavary AND Tony Z. Zhao AND Christopher Agia AND Rohan Baijal AND Mateo Guaman Castro AND Daphne Chen AND Qiuyu Chen AND Trinity Chung AND Jaimyn Drake AND Ethan Paul Foster AND Jensen Gao AND David Antonio Herrera AND Minho Heo AND Kyle Hsu AND Jiaheng Hu AND Donovon Jackson AND Charlotte Le AND Yunshuang Li AND Roy Lin AND Zehan Ma AND Abhiram Maddukuri AND Suvir Mirchandani AND Daniel Morton AND Tony Nguyen AND Abigail O'Neill AND Rosario Scalise AND Derick Seale AND Victor Son AND Stephen Tian AND Emi Tran AND Andrew E. Wang AND Yilin Wu AND Annie Xie AND Jingyun Yang AND Patrick Yin AND Yunchu Zhang AND Osbert Bastani AND Glen Berseth AND Jeannette Bohg AND Ken Goldberg AND Abhinav Gupta AND Abhishek Gupta AND Dinesh Jayaraman AND Joseph J Lim AND Jitendra Malik AND Roberto Martín-Martín AND Subramanian Ramamoorthy AND Dorsa Sadigh AND Shuran Song AND Jiajun Wu AND Michael C. Yip AND Yuke Zhu AND Thomas Kollar AND Sergey Levine AND Chelsea Finn}, 
    TITLE     = {{DROID}: A Large-Scale In-The-Wild Robot Manipulation Dataset}, 
    BOOKTITLE = rss, 
    YEAR      = {2024}
}

@inproceedings{robominddataset,
  author = {Wu, Kun and Hou, Chengkai and Liu, Jiaming and Che, Zhengping and Ju, Xiaozhu and Yang, Zhuqin and Li, Meng and Zhao, Yinuo and Xu, Zhiyuan and Yang, Guang and Fan, Shichao and Wang, Xinhua and Liao, Fei and Zhao, Zhen and Li, Guangyu and Jin, Zhao and Wang, Lecheng and Mao, Jilei and Liu, Ning and Ren, Pei and Zhang, Qiang and Lyu, Yaoxu and Liu, Mengzhen and Jingyang, He and Luo, Yulin and Gao, Zeyu and Li, Chenxuan and Gu, Chenyang and Fu, Yankai and Wu, Di and Wang, Xingyu and Chen, Sixiang and Wang, Zhenyu and An, Pengju and Qian, Siyuan and Zhang, Shanghang and Tang, Jian},
  booktitle=rss, 
  title={{RoboMIND}: Benchmark on Multi-embodiment Intelligence Normative Data for Robot Manipulation}, 
  year={2025}
}

@inproceedings{roboverse,
  title={{RoboVerse}: A Unified Platform, Benchmark and Dataset for Scalable and Generalizable Robot Learning},
  author={Haoran Geng and Feishi Wang and Songlin Wei and Yuyang Li and Bangjun Wang and Boshi An and Haozhe Lou and Charlie Tianyue Cheng and Peihao Li and Haozhe Chen and Yutong Liang and Yuxi Qian and Jiageng Mao and Weikang Wan and Yiran Geng and Mingtong Zhang and Jiangran Lyu and Siheng Zhao and Jiazhao Zhang and Chaoyi Xu and Jialiang Zhang and Chengyang Zhao and Haoran Lu and Yufei Ding and Ran Gong and Yuran Wang and Yuxuan Kuang and Ruihai Wu and Baoxiong Jia and Hao Dong and Siyuan Huang and Yue Wang and Jitendra Malik and Pieter Abbeel},
  booktitle=rss,
  year={2025}
}

@misc{stbibench,
  title = {{{ST-BiBench}}: {{Benchmarking Multi-Stream Multimodal Coordination}} in {{Bimanual Embodied Tasks}} for {{MLLMs}}},
  author = {Wu, Xin and Liang, Zhixuan and Ma, Yue and Hu, Mengkang and Qin, Zhiyuan and Li, Xiu},
  year = 2026,
  howpublished = {\url{https://arxiv.org/abs/2602.08392}}
}

@article{heo2023furniturebench,
  title={Furniturebench: Reproducible real-world benchmark for long-horizon complex manipulation},
  author={Heo, Minho and Lee, Youngwoon and Lee, Doohyun and Lim, Joseph J},
  journal=IJRR,
  year={2023}
}

@article{fmb,
  title = {{{FMB}}: {{A}} Functional Manipulation Benchmark for Generalizable Robotic Learning},
  author = {Luo, Jianlan and Xu, Charles and Liu, Fangchen and Tan, Liam and Lin, Zipeng and Wu, Jeffrey and Abbeel, Pieter and Levine, Sergey},
  year = 2025,
  journal = IJRR,
  volume = {44},
  number = {4}
}

@inproceedings{vlabench,
  title = {{{VLABench}}: {{A Large-Scale Benchmark}} for {{Language-Conditioned Robotics Manipulation}} with {{Long-Horizon Reasoning Tasks}}},
  booktitle = ICCV,
  author = {Zhang, Shiduo and Xu, Zhe and Liu, Peiju and Yu, Xiaopeng and Li, Yuan and Gao, Qinghui and Fei, Zhaoye and Yin, Zhangyue and Wu, Zuxuan and Jiang, Yu-Gang and Qiu, Xipeng},
  year = 2025
}

@misc{robotwin2,
      title={RoboTwin 2.0: A Scalable Data Generator and Benchmark with Strong Domain Randomization for Robust Bimanual Robotic Manipulation}, 
      author={Tianxing Chen and Zanxin Chen and Baijun Chen and Zijian Cai and Yibin Liu and Zixuan Li and Qiwei Liang and Xianliang Lin and Yiheng Ge and Zhenyu Gu and Weiliang Deng and Yubin Guo and Tian Nian and Xuanbing Xie and Qiangyu Chen and Kailun Su and Tianling Xu and Guodong Liu and Mengkang Hu and Huan-ang Gao and Kaixuan Wang and Zhixuan Liang and Yusen Qin and Xiaokang Yang and Ping Luo and Yao Mu},
      year={2025},
      howpublished={\url{https://arxiv.org/abs/2506.18088}}
}

@misc{bicoord,
      title={BiCoord: A Bimanual Manipulation Benchmark towards Long-Horizon Spatial-Temporal Coordination}, 
      author={Xingyu Peng and Chen Gao and Liankai Jin and Annan Li and Si Liu},
      year={2026},
      howpublished={\url{https://arxiv.org/abs/2604.05831}}
}

@INPROCEEDINGS{rlbench2,
  author={Grotz, Markus and Shridhar, Mohit and Chao, Yu-Wei and Asfour, Tamim and Fox, Dieter},
  booktitle=ICRA, 
  title={TWIN: Two-handed Intelligent Benchmark for Bimanual Manipulation}, 
  year={2025},
  doi={10.1109/ICRA55743.2025.11128527}}

@misc{rcs,
  title={{Robot Control Stack: A Lean Ecosystem for Robot Learning at Scale}}, 
  author={Tobias J{\"u}lg and Pierre Krack and Seongjin Bien and Yannik Blei and Khaled Gamal and Ken Nakahara and Johannes Hechtl and Roberto Calandra and Wolfram Burgard and Florian Walter},
  year={2025},
  howpublished = {\url{https://arxiv.org/abs/2509.14932}}
}

@misc{vlagents,
      title={{VLAgents: A Policy Server for Efficient VLA Inference}}, 
      author={Tobias J{\"u}lg and Khaled Gamal and Nisarga Nilavadi and Pierre Krack and Seongjin Bien and Michael Krawez and Florian Walter and Wolfram Burgard},
      year={2026},
      howpublished={\url{https://arxiv.org/abs/2601.11250}}
}

@ARTICLE{krebs,
  author={Krebs, Franziska and Asfour, Tamim},
  journal=ieeeral, 
  title={A Bimanual Manipulation Taxonomy}, 
  year={2022},
  volume={7},
  number={4},
  doi={10.1109/LRA.2022.3196158}}

@InProceedings{iris,
  title = 	 {{IRIS}: An Immersive Robot Interaction System},
  author =       {Jiang, Xinkai and Yuan, Qihao and Dincer, Enes Ulas and Zhou, Hongyi and Li, Ge and Li, Xueyin and Jia, Xiaogang and Schnizer, Timo and Schreiber, Nicolas and Liao, Weiran and Haag, Julius and Li, Kailai and Neumann, Gerhard and Lioutikov, Rudolf},
  booktitle = 	 CORL,
  year = 	 {2025},
  volume = 	 {305}
}

@software{menagerie2022github,
  author = {Zakka, Kevin and Tassa, Yuval and {MuJoCo Menagerie Contributors}},
  title = {{MuJoCo Menagerie: A collection of high-quality simulation models for MuJoCo}},
  url = {http://github.com/google-deepmind/mujoco_menagerie},
  year = {2022}
}

@book{sutton1998reinforcement,
  title={Reinforcement learning: An introduction},
  author={Sutton, Richard S and Barto, Andrew G and others},
  volume={1},
  year={1998},
  publisher={MIT press Cambridge}
}

@InProceedings{act,
  title={Learning Fine-Grained Bimanual Manipulation with Low-Cost Hardware},
  author={Zhao, Tony and Kumar, Vikash and Levine, Sergey and Finn, Chelsea},
  booktitle=rss,
  year={2023}
}

@inproceedings{piofive,
      title={$\pi_{0.5}$: a Vision-Language-Action Model with Open-World Generalization}, 
      author={Physical Intelligence and Kevin Black and Noah Brown and James Darpinian and Karan Dhabalia and Danny Driess and Adnan Esmail and Michael Equi and Chelsea Finn and Niccolo Fusai and Manuel Y. Galliker and Dibya Ghosh and Lachy Groom and Karol Hausman and Brian Ichter and Szymon Jakubczak and Tim Jones and Liyiming Ke and Devin LeBlanc and Sergey Levine and Adrian Li-Bell and Mohith Mothukuri and Suraj Nair and Karl Pertsch and Allen Z. Ren and Lucy Xiaoyang Shi and Laura Smith and Jost Tobias Springenberg and Kyle Stachowicz and James Tanner and Quan Vuong and Homer Walke and Anna Walling and Haohuan Wang and Lili Yu and Ury Zhilinsky},
      booktitle=CORL,
      year={2025}
}

@misc{xvla,
      title={{X-VLA}: Soft-Prompted Transformer as Scalable Cross-Embodiment Vision-Language-Action Model}, 
      author={Jinliang Zheng and Jianxiong Li and Zhihao Wang and Dongxiu Liu and Xirui Kang and Yuchun Feng and Yinan Zheng and Jiayin Zou and Yilun Chen and Jia Zeng and Ya-Qin Zhang and Jiangmiao Pang and Jingjing Liu and Tai Wang and Xianyuan Zhan},
      year={2025},
      howpublished={\url{https://arxiv.org/abs/2510.10274}}
}

@misc{franka_docs,
  author       = {{Franka Robotics GmbH}},
  title        = {Franka Documentation Portal},
  year         = {2026},
  howpublished = {\url{https://www.franka.de/documents}}
}

@misc{roboeval,
  title         = {{RoboEval}: Where Robotic Manipulation Meets Structured and Scalable Evaluation},
  author        = {Wang, Yi Ru and Ung, Carter and Tan, Christopher and Tannert, Grant and Duan, Jiafei and Li, Josephine and Le, Anh and Oswal, Rishabh and Grotz, Markus and Pumacay, Wilbert and Deng, Yuquan and Krishna, Ranjay and Fox, Dieter and Srinivasa, Siddhartha},
  year          = {2025},
  eprint        = {2507.00435},
  archivePrefix = {arXiv},
  primaryClass  = {cs.RO}
}

@misc{kressgazit2024empirical,
  title         = {Robot Learning as an Empirical Science: Best Practices for Policy Evaluation},
  author        = {Kress-Gazit, Hadas and Hashimoto, Kunimatsu and Kuppuswamy, Naveen and Shah, Paarth and Horgan, Phoebe and Richardson, Gordon and Feng, Siyuan and Burchfiel, Benjamin},
  year          = {2024},
  eprint        = {2409.09491},
  archivePrefix = {arXiv},
  primaryClass  = {cs.RO}
}

@article{molmospaces,
  title={MolmoSpaces: A Large-Scale Open Ecosystem for Robot Navigation and Manipulation},
  author={Kim, Yejin and Pumacay, Wilbert and Rayyan, Omar and Argus, Max and Han, Winson and VanderBilt, Eli and Salvador, Jordi and Deshpande, Abhay and Hendrix, Rose and Jauhri, Snehal and Liu, Shuo and Shafiullah, Nur Muhammad Mahi and Guru, Maya and Eftekhar, Ainaz and Farley, Karen and Clay, Donovan and Duan, Jiafei and Guru, Arjun and Wolters, Piper and Herrasti, Alvaro and Lee, Ying-Chun and Chalvatzaki, Georgia and Cui, Yuchen and Farhadi, Ali and Fox, Dieter and Krishna, Ranjay},
  journal={arXiv preprint arXiv:2602.11337},
  year={2026}
}

\clearpage
\appendix

\section{Extended Results}
\label{app:results}

\subsection{Task Progress over Time}
\begin{figure}[H]
	\centering
	\includegraphics[width=0.9\linewidth]{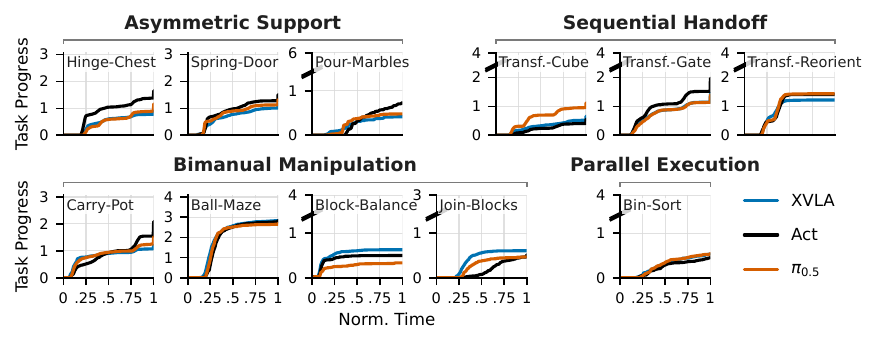}
	\caption{Average task progress over normalized time across all rollouts in simulation.}
    \label{fig:avg_task_progress}
\end{figure}

\autoref{fig:avg_task_progress} shows average task progress over normalized time across 300 simulation rollouts for each task. Curves that rise earlier indicate policies that reach early stages faster; plateaus reveal stages where policies commonly stall. Comparing these curves highlights differences in data efficiency, the speed of acquiring coordination-relevant milestones, and where policies tend to lose progress during execution.

\subsection{Failure-Stage Distributions}
\begin{figure}[H]
    \centering
    \includegraphics[width=0.7\linewidth]{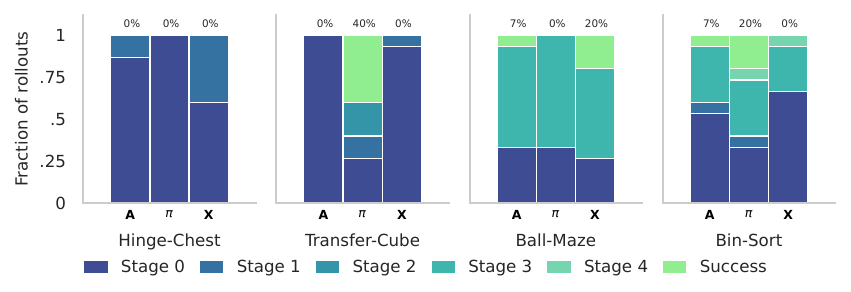}
    \caption{Fraction of real-world rollouts that ended in a given stage. Green means fraction of successful episodes annotated with the success rate. Policy names are abbreviated as follows: \act{} as ``A'', \piofive{} as ``$\pi$'' and \xvla{} as ``X''.}
    \label{fig:stages:real}
\end{figure}

In \autoref{fig:stages:real}, each bar shows the fraction of real-world rollouts ($N=15$ each) that terminated in a particular stage, with the green segment denoting successful episodes. These distributions make it easy to identify dominant failure stages (e.g., grasp acquisition versus later manipulation) and to compare failure-mode patterns across policies and tasks, helping to identify which subtasks most limit real-world performance.

\subsection{Pass-Rate Profiles}
\begin{figure}[H]
	\centering
	\includegraphics[width=\linewidth]{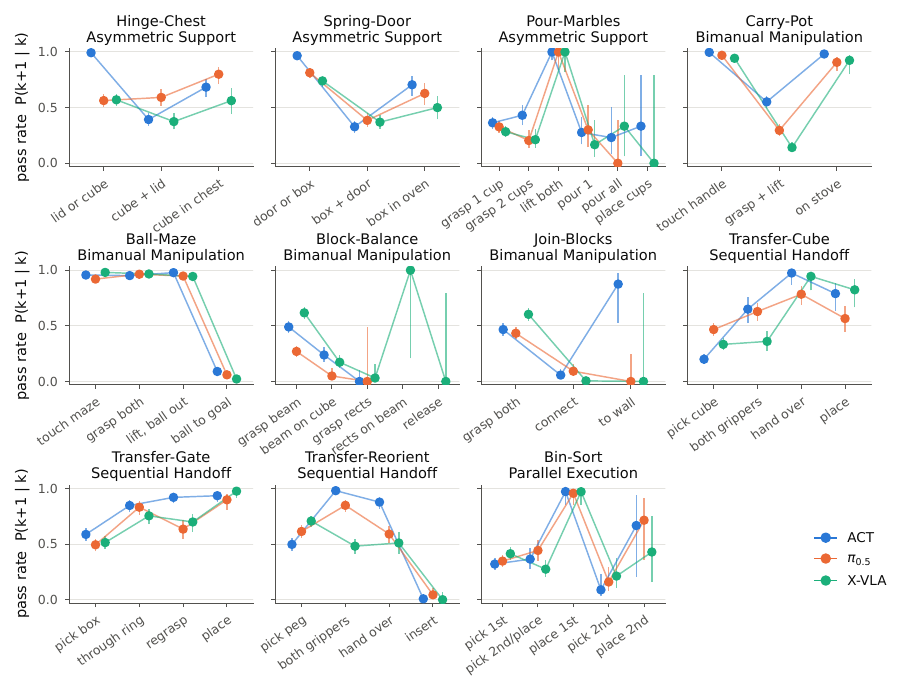}
	\caption{Stage-conditional pass rates $\passk$ with Wilson 95\% intervals for all eleven tasks in simulation (300 rollouts per policy and task). The $x$-axis names the sub-instruction to leave stage $k$; a series ends where no episode reached the next stage.}
	\label{fig:passrates:sim}
\end{figure}

\autoref{fig:passrates:sim} gives the stage-conditional pass rates for all eleven simulation tasks; \autoref{tab:sim} in the main text lists only the stage with the lowest pass rate from each of these profiles. A series ends at the first stage that no rollout reached, so short series mark tasks where all policies fail early.

\subsection{Real-World Results}
\autoref{tab:real} reports the real-world evaluation of the policies trained on real data only, with Wilson intervals and the bottleneck stage, and complements the simulation-only \autoref{tab:sim}; the comparison with mixed training is given in \autoref{tab:mix} in the main text.

\begin{table}[H]
\centering
\caption{Real-world results, 15 rollouts per cell, policies trained on real data only: success rate (SR, \%) with Wilson 95\% interval, $p_\text{final}$, and the bottleneck stage with its pass rate $\passk$.}
\label{tab:real}
\resizebox{\columnwidth}{!}{%
\begin{tabular}{lrclrclrcl}
\toprule
Task & \multicolumn{3}{c}{\act} & \multicolumn{3}{c}{\piofive} & \multicolumn{3}{c}{\xvla} \\
\cmidrule(lr){2-4}\cmidrule(lr){5-7}\cmidrule(lr){8-10}
 & SR [CI] & $p_\text{final}$ & bottleneck stage ($\rho_k$) & SR [CI] & $p_\text{final}$ & bottleneck stage ($\rho_k$) & SR [CI] & $p_\text{final}$ & bottleneck stage ($\rho_k$) \\
\midrule
\hingechest & 0\,\textcolor{gray}{\scriptsize[0,20]} & \textcolor{gray}{0.04} & cube + lid \textcolor{gray}{\scriptsize(0.00)} & 0\,\textcolor{gray}{\scriptsize[0,20]} & \textcolor{gray}{0.00} & lid or cube \textcolor{gray}{\scriptsize(0.00)} & 0\,\textcolor{gray}{\scriptsize[0,20]} & \textcolor{gray}{0.13} & cube + lid \textcolor{gray}{\scriptsize(0.00)} \\
\ballmaze & 7\,\textcolor{gray}{\scriptsize[1,30]} & \textcolor{gray}{0.52} & ball to goal \textcolor{gray}{\scriptsize(0.10)} & 0\,\textcolor{gray}{\scriptsize[0,20]} & \textcolor{gray}{0.50} & ball to goal \textcolor{gray}{\scriptsize(0.00)} & 20\,\textcolor{gray}{\scriptsize[7,45]} & \textcolor{gray}{0.60} & ball to goal \textcolor{gray}{\scriptsize(0.27)} \\
\transfercube & 0\,\textcolor{gray}{\scriptsize[0,20]} & \textcolor{gray}{0.00} & pick cube \textcolor{gray}{\scriptsize(0.00)} & 40\,\textcolor{gray}{\scriptsize[20,64]} & \textcolor{gray}{0.53} & hand over \textcolor{gray}{\scriptsize(0.67)} & 0\,\textcolor{gray}{\scriptsize[0,20]} & \textcolor{gray}{0.02} & both grippers \textcolor{gray}{\scriptsize(0.00)} \\
\binsort & 7\,\textcolor{gray}{\scriptsize[1,30]} & \textcolor{gray}{0.28} & pick 2nd \textcolor{gray}{\scriptsize(0.17)} & 20\,\textcolor{gray}{\scriptsize[7,45]} & \textcolor{gray}{0.47} & pick 2nd \textcolor{gray}{\scriptsize(0.44)} & 0\,\textcolor{gray}{\scriptsize[0,20]} & \textcolor{gray}{0.21} & place 2nd \textcolor{gray}{\scriptsize(0.00)} \\
\bottomrule
\end{tabular}

}
\end{table}

\section{Training Details}
\label{app:training}
All policies are trained with the LeRobot framework~\cite{lerobot} on a single node using Hugging Face Accelerate with bf16 mixed precision; no image augmentation is applied.
The teleoperated episodes are converted to the LeRobot dataset format with the three camera streams (head, left wrist, right wrist) resized to $224\times224$ pixels.
Observations are the joint positions of both arms and both grippers, and actions are absolute joint targets for both arms, clipped to the joint limits of the arms, with binarized gripper commands.
Mixed datasets are built by merging the per-domain datasets with the LeRobot dataset tools.
\autoref{tab:training} lists the hyperparameters; values that are not set explicitly in our launch scripts are the LeRobot (v0.5.1) defaults of the respective policy and are listed because they may change between versions. All runs use the default training seed, one observation step, and no temporal ensembling for \act{}.
\act{} is trained from scratch per task with an ImageNet-pretrained ResNet-18 backbone, whereas \piofive{} and \xvla{} are fully fine-tuned from the public checkpoints \texttt{lerobot/pi05\_base} and \texttt{lerobot/xvla-base}, i.e., the vision and language encoders are not frozen.
For \xvla{}, the same domain embedding is used for simulation and real-world data (cf.\ \autoref{sec:results}); its ``auto'' action space zero-pads the joint action vector to the model's 20 dimensions and trains with a plain MSE loss on all real dimensions including the grippers. Normalization and chunk size follow the defaults of the respective policy, for \xvla{} those of the base checkpoint.
At inference, every policy predicts an action chunk and executes 30 actions, corresponding to one second open loop, before it is prompted again.

\begin{table}[H]
\centering
\caption{Training hyperparameters. Batch size is per GPU; the effective batch size is given in parentheses.}
\label{tab:training}
\resizebox{\columnwidth}{!}{%
\begin{tabular}{lccc}
\toprule
 & \act{} & \piofive{} & \xvla{} \\
\midrule
Initialization & from scratch (ImageNet ResNet-18) & \texttt{lerobot/pi05\_base} & \texttt{lerobot/xvla-base} \\
Trainable parameters & all & all (VLM incl.\ vision encoder, action expert) & all (encoders, transformer, soft prompts) \\
Training regime & one policy per task & one policy for all tasks & one policy for all tasks \\
Training steps & 150\,000 & 40\,000 & 40\,000 \\
Batch size & 64 (64) & 16 (64) & 48 (192) \\
Optimizer & AdamW, $\beta=(0.9, 0.999)$ & AdamW, $\beta=(0.9, 0.95)$ & AdamW, $\beta=(0.9, 0.99)$ \\
Learning rate & $1\times10^{-5}$, constant & $2.5\times10^{-5}$, cosine decay & $1\times10^{-4}$, cosine decay \\
Warmup / decay steps & -- & 1\,000 / 30\,000 (to $2.5\times10^{-6}$) & 1\,000 / 30\,000 (to $2.5\times10^{-6}$) \\
Weight decay & $10^{-4}$ & 0.01 & 0 \\
Gradient clipping & 10.0 & 1.0 & 10.0 \\
Action chunk size & 100 & 50 & 30 \\
Normalization (state / action) & mean-std / mean-std & quantiles / quantiles & none / mean-std \\
Inference flow / denoising steps & -- & 10 & 10 \\
Precision & bf16 & bf16 & bf16 \\
GPUs & 1$\times$ H100 & 4$\times$ H200 & 4$\times$ H100 \\
\bottomrule
\end{tabular}%
}
\end{table}

\section{Visual Randomization}
\begin{figure}[h]
	\centering
	\includegraphics[width=\linewidth]{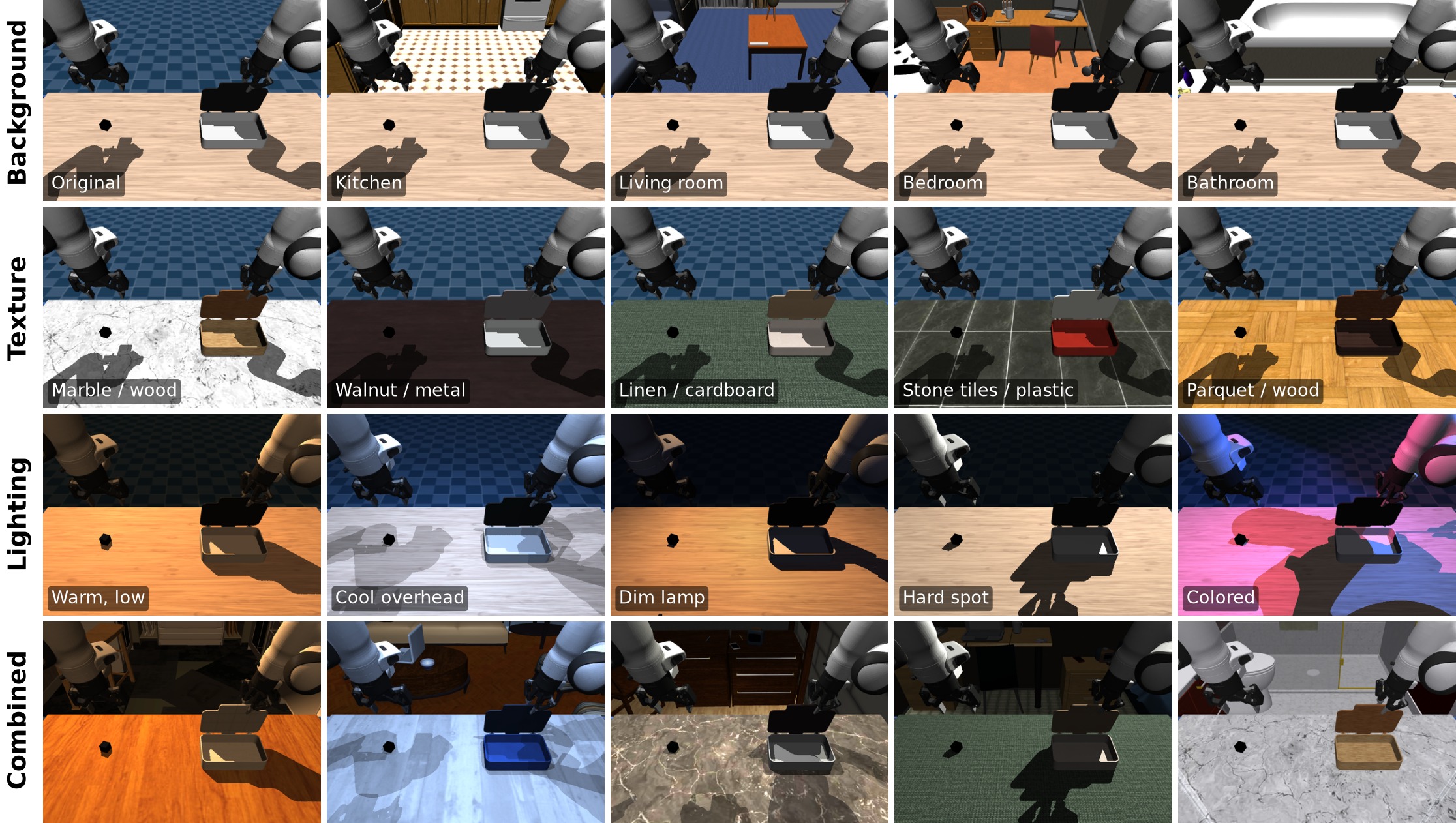}
	\caption{Visual randomization of one \hingechest{} frame from the head camera. Row 1: default scene (left) and MolmoSpaces~\cite{molmospaces} backgrounds. Row 2: table and chest textures. Row 3: lighting. Row 4: all combined.}
    \label{fig:visual_ablations}
\end{figure}
The benchmark includes a replayer that restores the MuJoCo state of all objects from a recorded teleoperation episode and re-records the camera images. Since replay depends only on the recorded states, the scene description can be modified beforehand, which allows re-rendering any simulation episode under visual randomization without additional teleoperation data. \autoref{fig:visual_ablations} demonstrates this for three factors: backgrounds, by placing the setup in hand-crafted MolmoSpaces~\cite{molmospaces} scenes; textures, by replacing table and object materials with realistic surface textures; and lighting. Objects referenced in the language instruction, such as the cube in \hingechest{}, keep their appearance. Our experiments deliberately use the default visuals to keep simulation and real-world results comparable; training and evaluating policies under such visual shifts is left for future work.
\section{3D Printing and Assembly}

All assets required for the real-world experiments are designed to be 3D-printable. In our experiments, the assets were printed on a Bambulab X1C printer. The assets are directly available in the project's GitHub repository.
The Hinge Chest requires an additional assembly step to attach the lid to the chest. In our experiments, we attached the parts using a pair of M3$\times$2cm screws and nuts.

\begin{landscape}
\section{Tasks}
\label{app:tasks}

\small
\setlength{\tabcolsep}{4pt}
\renewcommand{\arraystretch}{1.15}

\setlength{\LTcapwidth}{\dimexpr
3cm + 1.8cm + 5cm + 1.8cm + 0.8cm + 0.8cm + 1cm + 4cm + 2cm
+ 18\tabcolsep
\relax}

\begin{longtable}[t]{ m{3cm} m{1.8cm} m{5cm} m{1.8cm} m{0.8cm} m{0.8cm} m{1cm} m{4cm} m{2cm} }

\caption{Overview of all benchmark tasks grouped into the categories of sequential handover, parallel execution, asymmetric support, and bimanual manipulation. The table summarizes whether tasks were replicated in a real-world setup, whether they are suitable for 3D printing, the 99th percentile episode length in steps as an indicator of task horizon, and the principal subtasks involved. Additionally, the employed randomization strategies are reported.}
\label{tab:task_overview_full} \\
\toprule
\textbf{Scene} &
\textbf{Task} &
\textbf{Description} &
\textbf{Category} &
\textbf{3D Printable} &
\textbf{Real-World} &
\textbf{Steps} &
\textbf{Key Subtasks} &
\textbf{Randomization} \\
\midrule
\endfirsthead

\toprule
\textbf{Scene} &
\textbf{Task} &
\textbf{Description} &
\textbf{Category} &
\textbf{3D Printable} &
\textbf{Real-World} &
\textbf{Steps} &
\textbf{Stages} &
\textbf{Randomization} \\
\midrule
\endhead

\includegraphics[width=3cm]{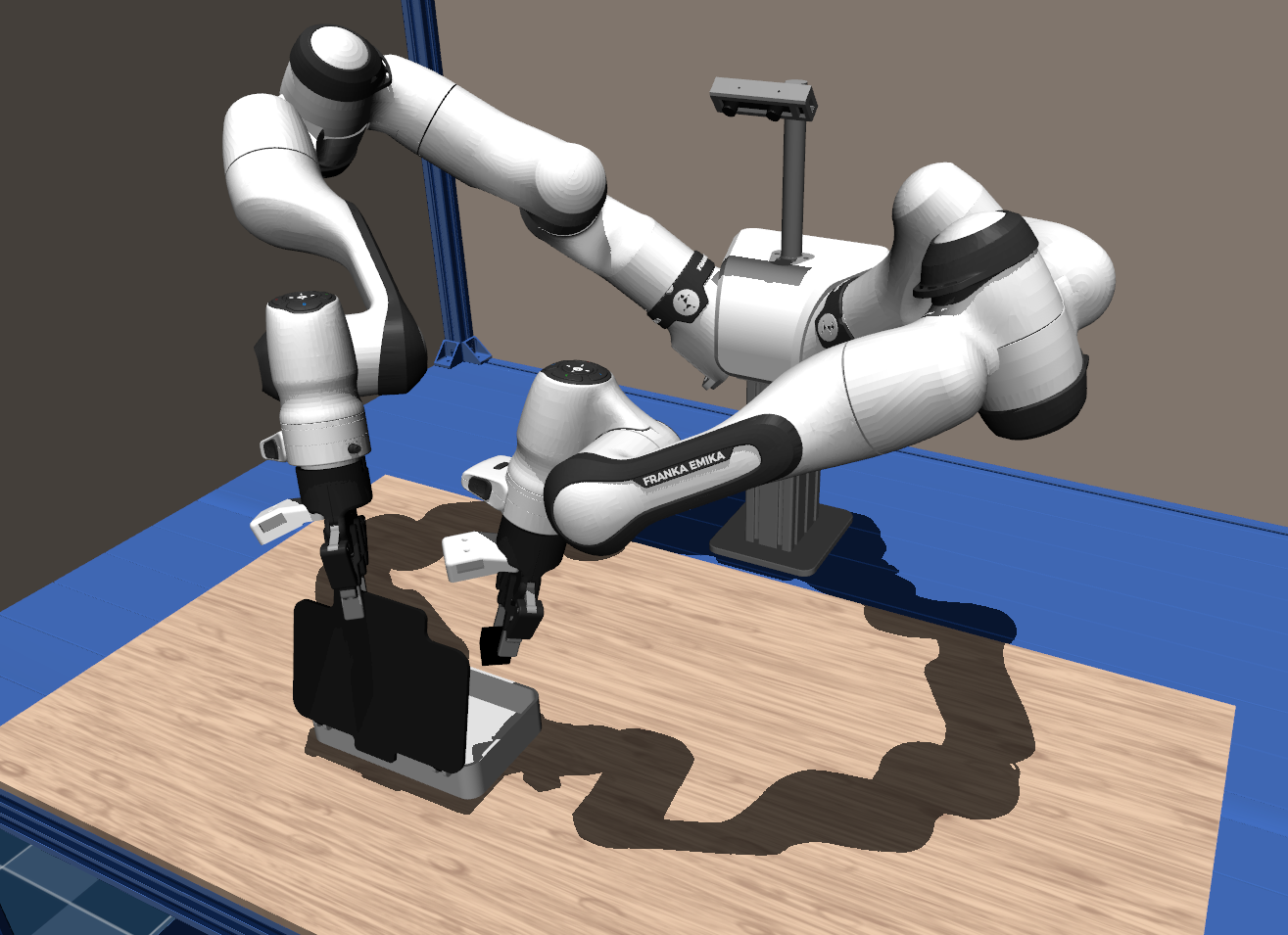} & \textbf{\hingechest} & Holding the lid of a small chest open while inserting a cube. One arm must hold the lid while the other inserts the cube. & Asymmetric Support & \checkmark & \checkmark & $441$ & open the chest OR pick up the cube, pick up the cube AND open the chest, place the cube inside the chest & cube position, chest position\\

\includegraphics[width=3cm]{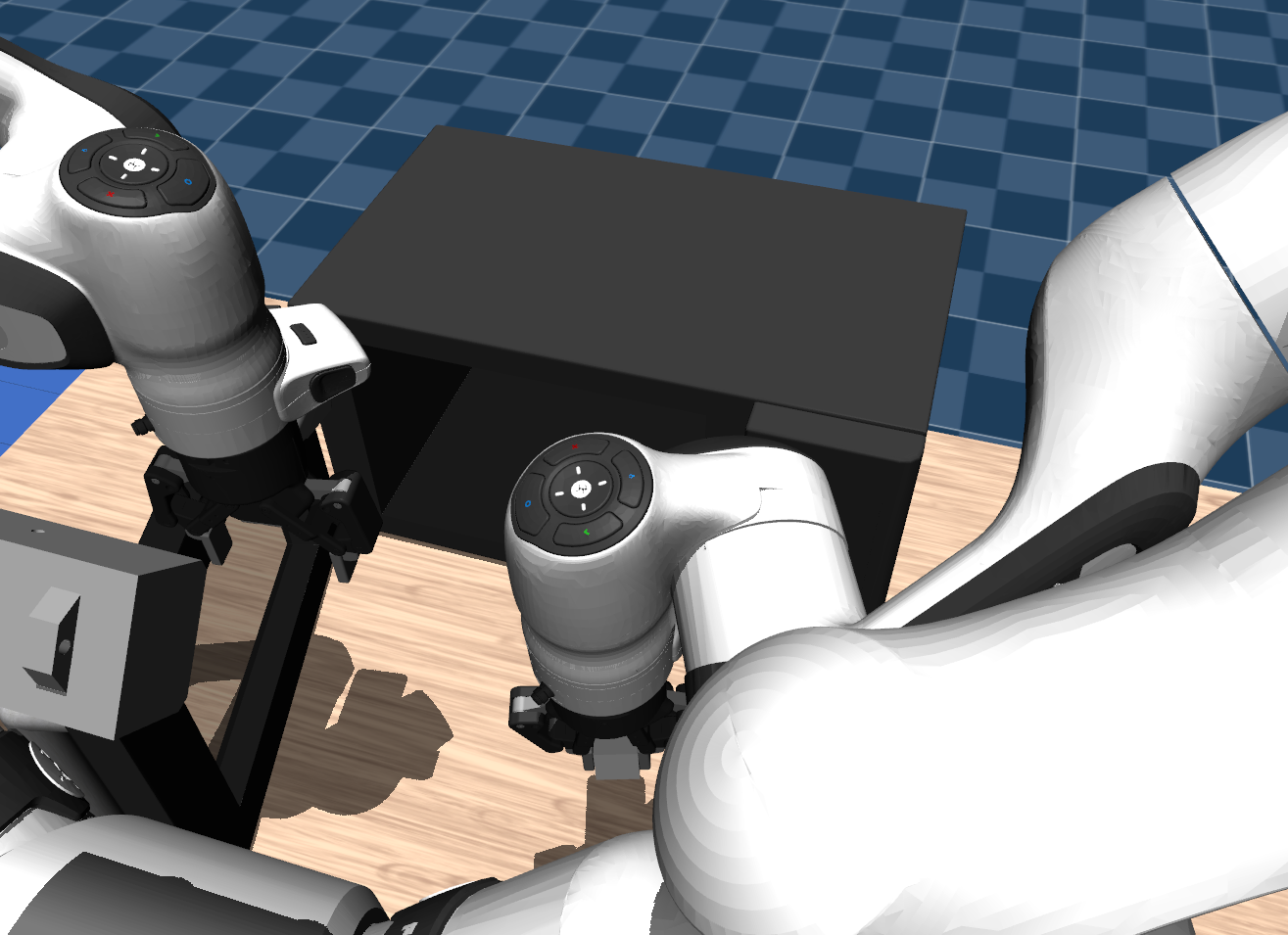} & \textbf{\springdoor} & A spring-loaded microwave door requires one arm to hold it open while the other inserts a box. & Asymmetric Support & $\times$ & $\times$ & $807$ & open the microwave OR pick up the box, pick up the box AND open the microwave, place the box inside the microwave & box position, microwave position \\

\includegraphics[width=3cm]{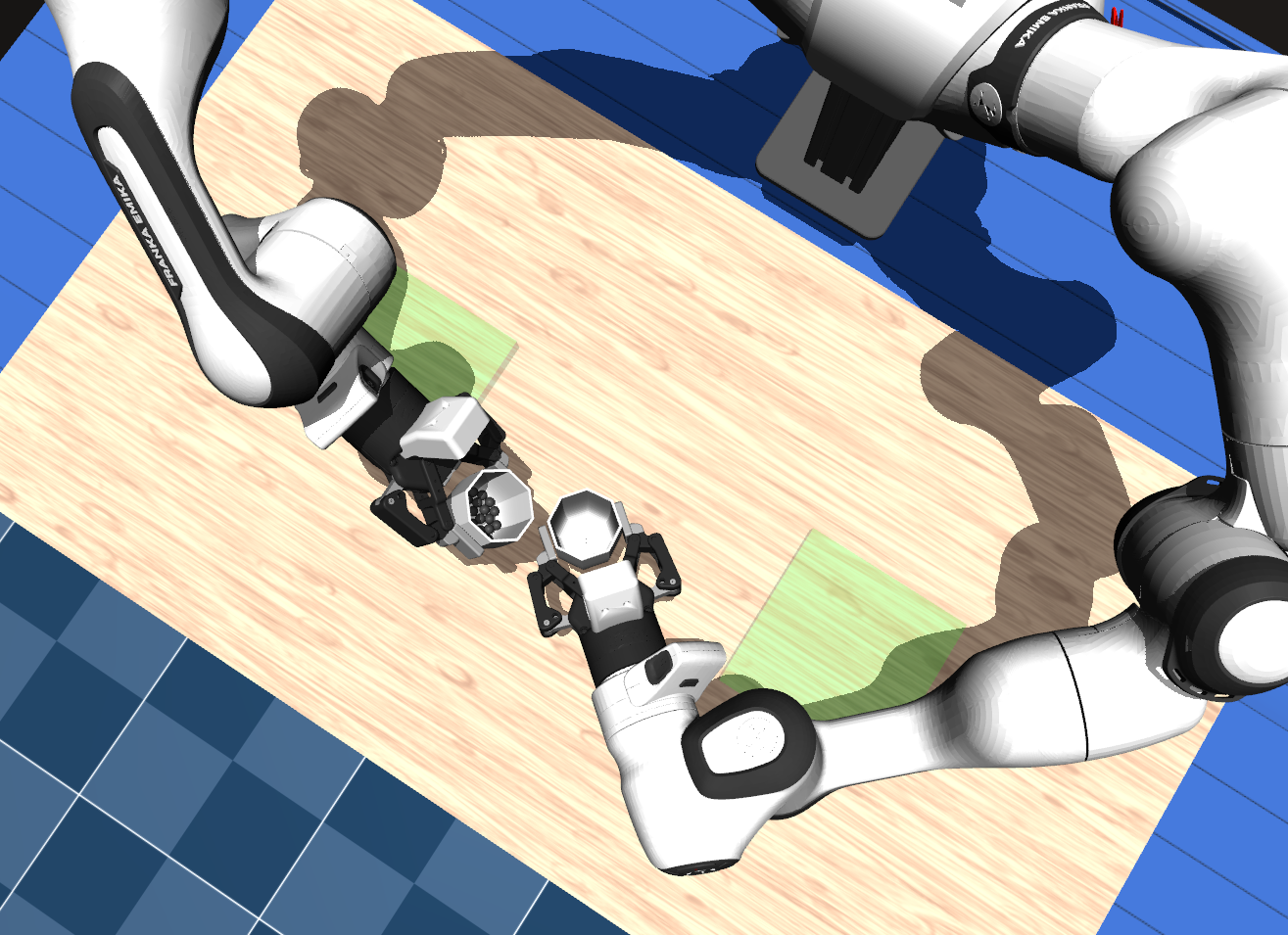} & \textbf{\pourmarbles} & Two cups, one containing marbles. Both cups must be picked up, and the marbles must be poured into the other cup before both cups are placed back. & Asymmetric Support & $\times$ & $\times$ & $442$ & grasp at least one cup, grasp both cups, lift both cups, pour at least one marble into the target cup, pour all marbles into the target cup, place both cups & cup position, which cup contains marbles \\

\includegraphics[width=3cm]{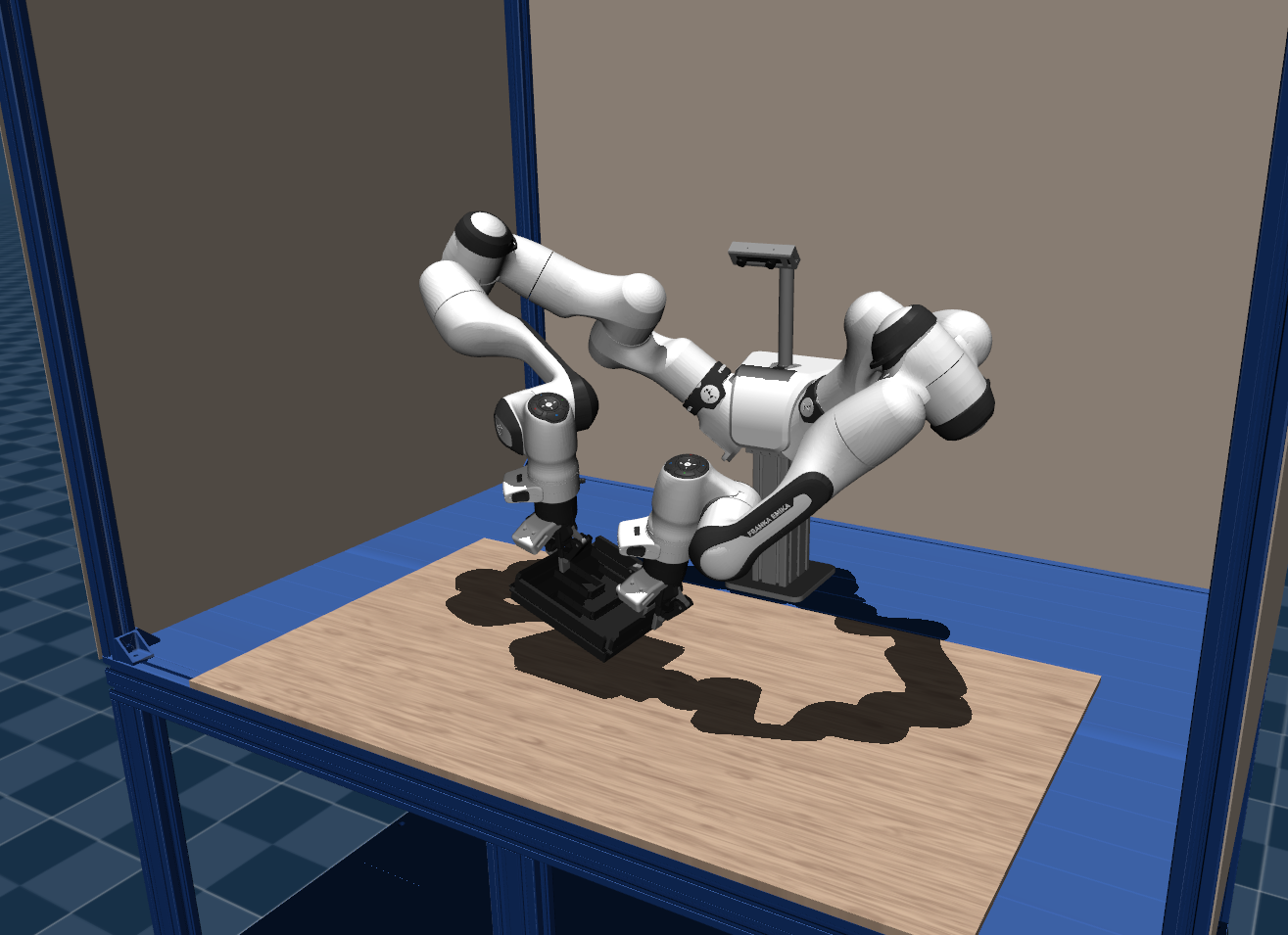} & \textbf{\ballmaze} & Pick up a maze board with both arms and tilt it so a ball rolls into a target region. & Bimanual Manipulation & \checkmark & \checkmark & $350$ & make contact with the maze, grasp the maze with both arms, lift the maze and move the ball out of the start area, guide the ball from the start area toward the goal & board position, one out of 10 different boards selected\\

\includegraphics[width=3cm]{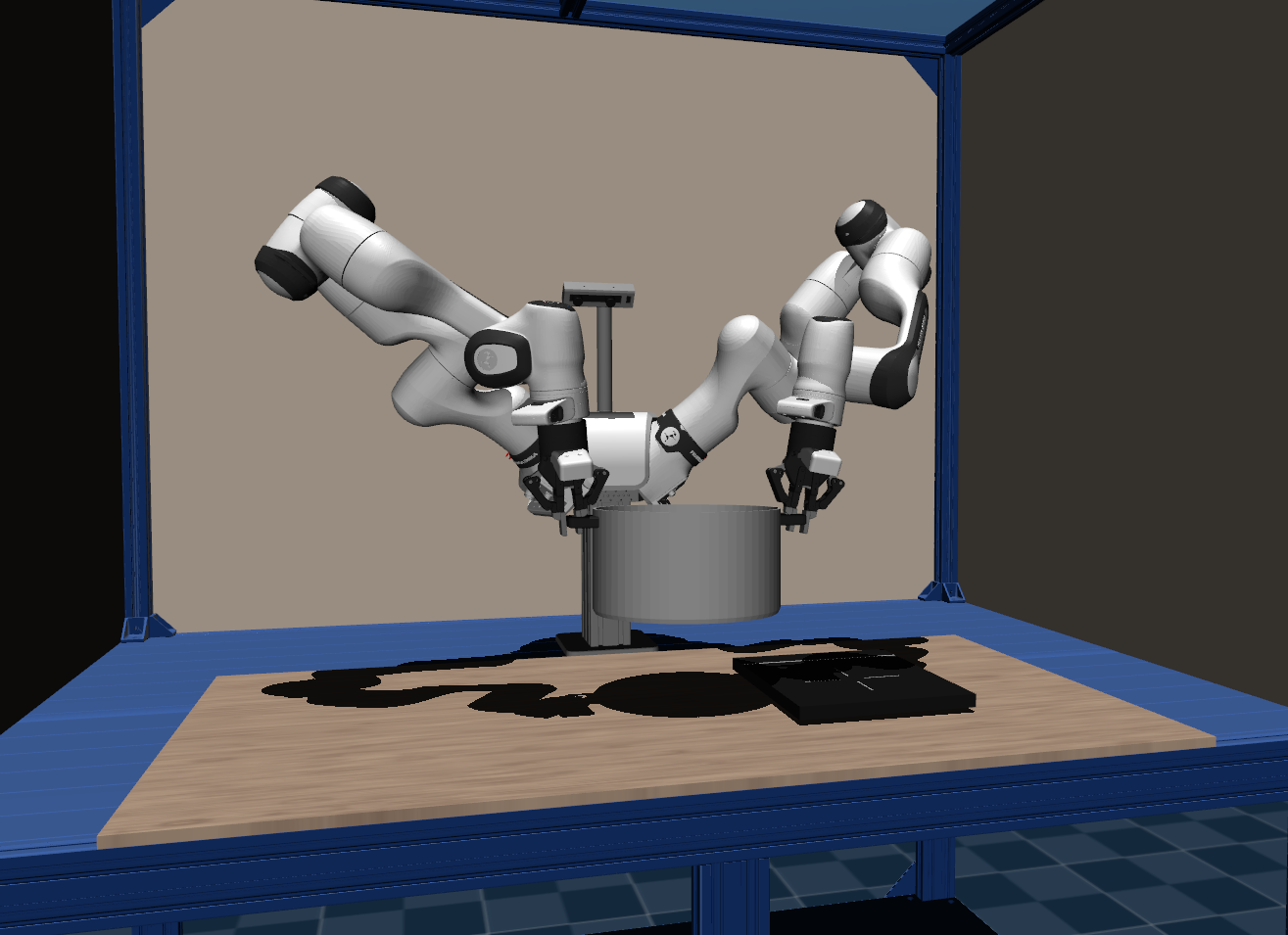} & \textbf{\carrypot} & Carry a pot using both side handles and place it on a stove. Both arms are needed to lift the pot. & Bimanual Manipulation & $\times$ & $\times$ & $450$ & make contact with a pot handle, grasp the pot by both handles and lift it, place the pot onto the stove & stove and pot positions within one half of the table each \\

\includegraphics[width=3cm]{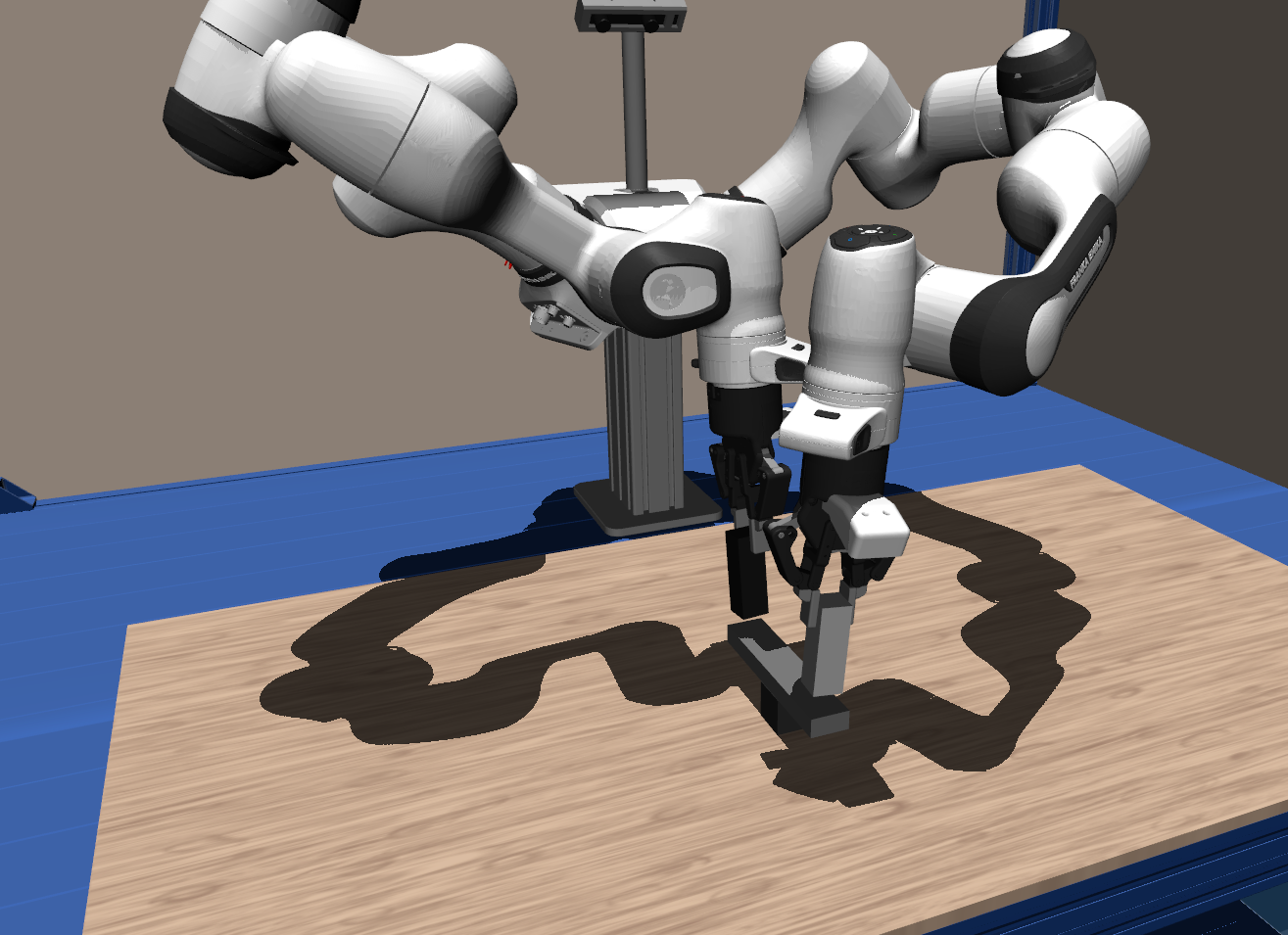} & \textbf{\blockbalance} & Consists of a red support cube, a beam, and two additional rectangular blocks. The beam needs to be placed on the cube, and then both blocks need to be placed on the beam simultaneously. & Bimanual Manipulation & \checkmark & $\times$ & $766$ & grasp the beam with one arm, place the beam on the small cube, grasp both rectangles, place both rectangles onto the beam, release and retract while keeping the beam balanced & position of all blocks \\

\includegraphics[width=3cm]{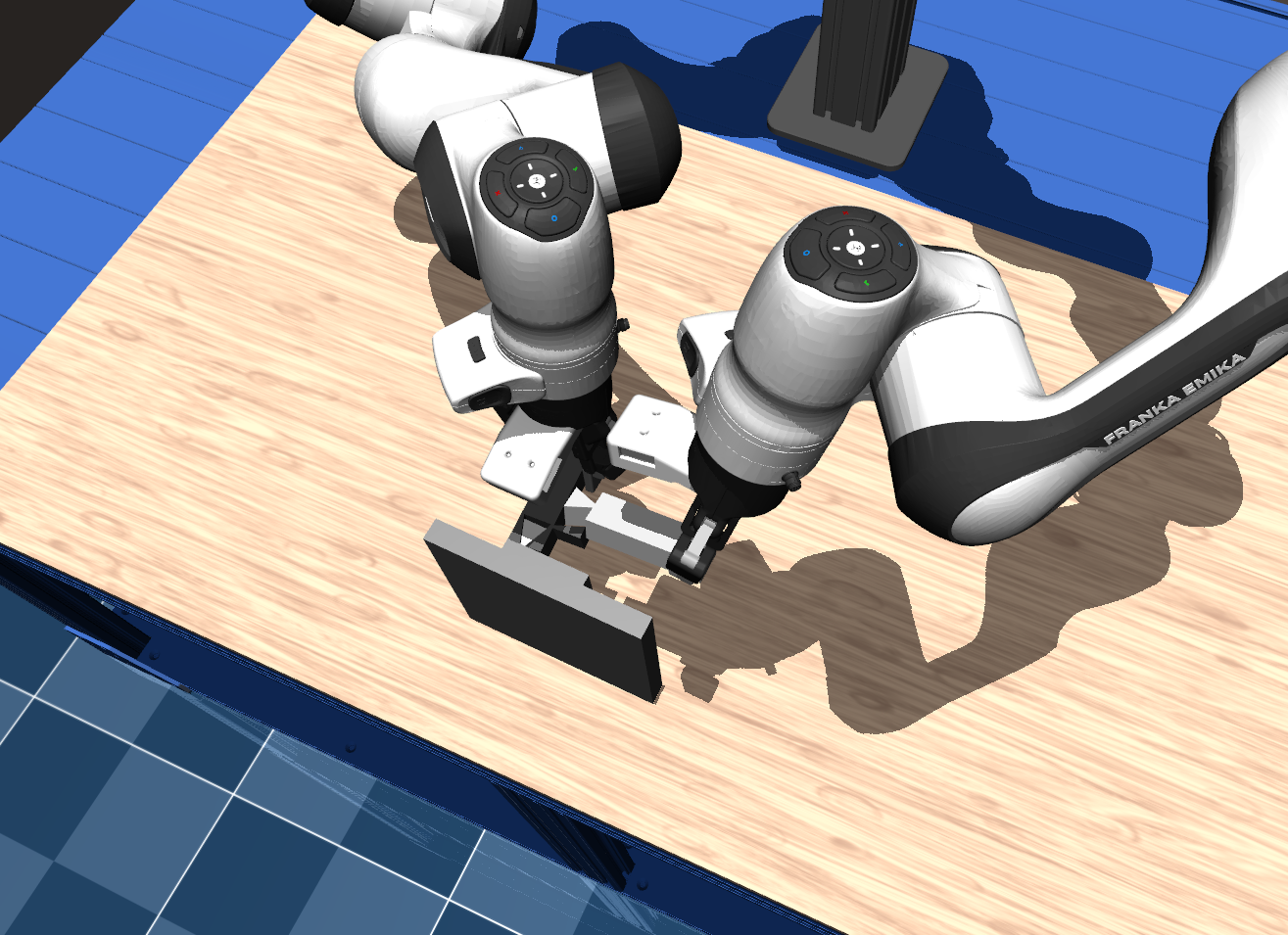} & \textbf{\joinblocks} & Connect two movable blocks together and then attach them to a peg on a third stationary block. & Bimanual Manipulation & \checkmark & $\times$ & $829$ & approach and grasp both blocks, connect the two blocks, connect the assembled blocks to the wall & position of both movable blocks, each on one half of the table \\

\includegraphics[width=3cm]{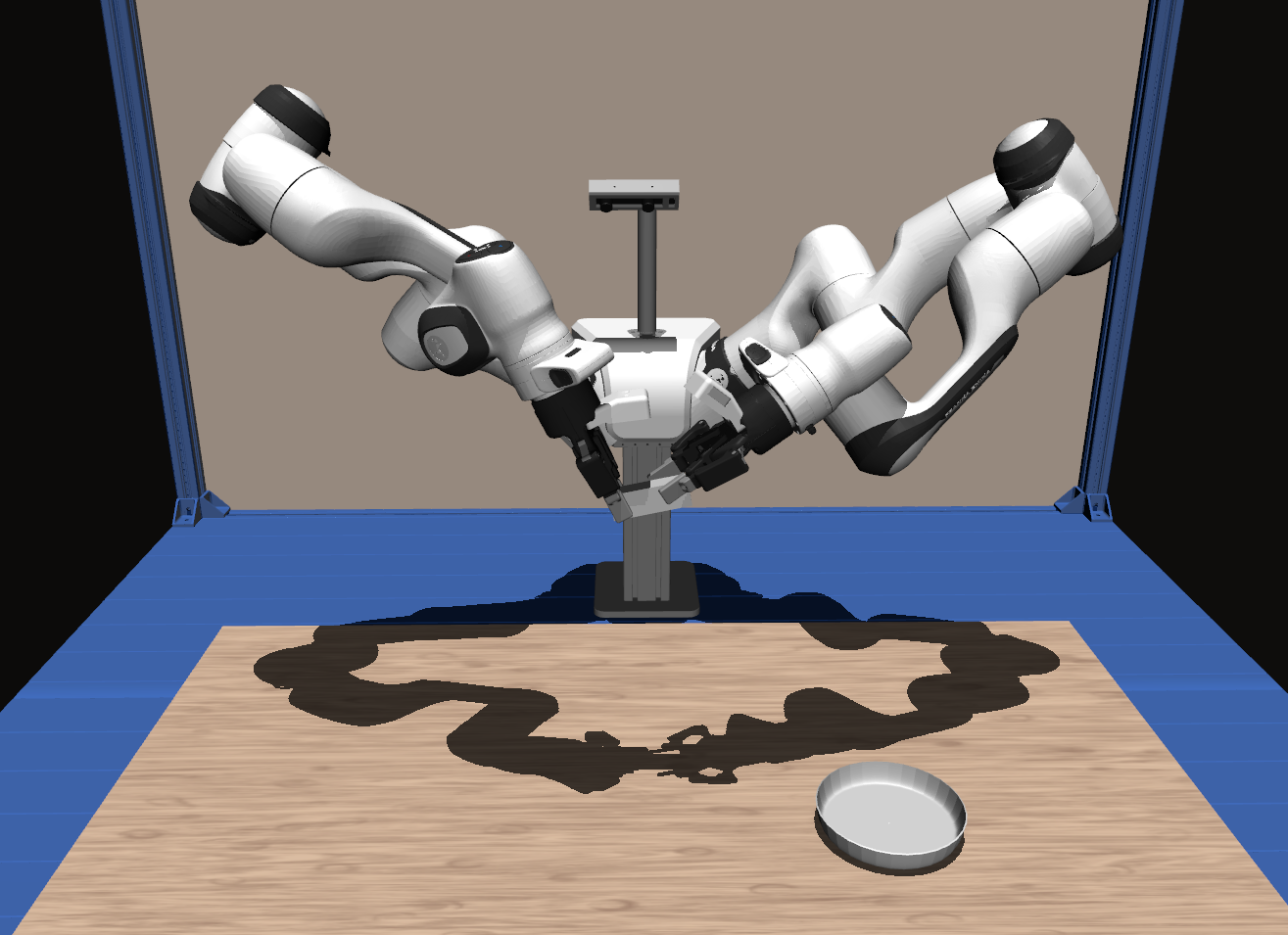} & \textbf{\transfercube} & Hand over a cube between arms before placing it into a bowl. & Sequential Handoff & \checkmark & \checkmark & $441$ & pick up the cube, bring both grippers into contact with the cube, transfer the cube to the other gripper, place the cube correctly and release it & cube and bowl positions, each on one half of the table \\

\includegraphics[width=3cm]{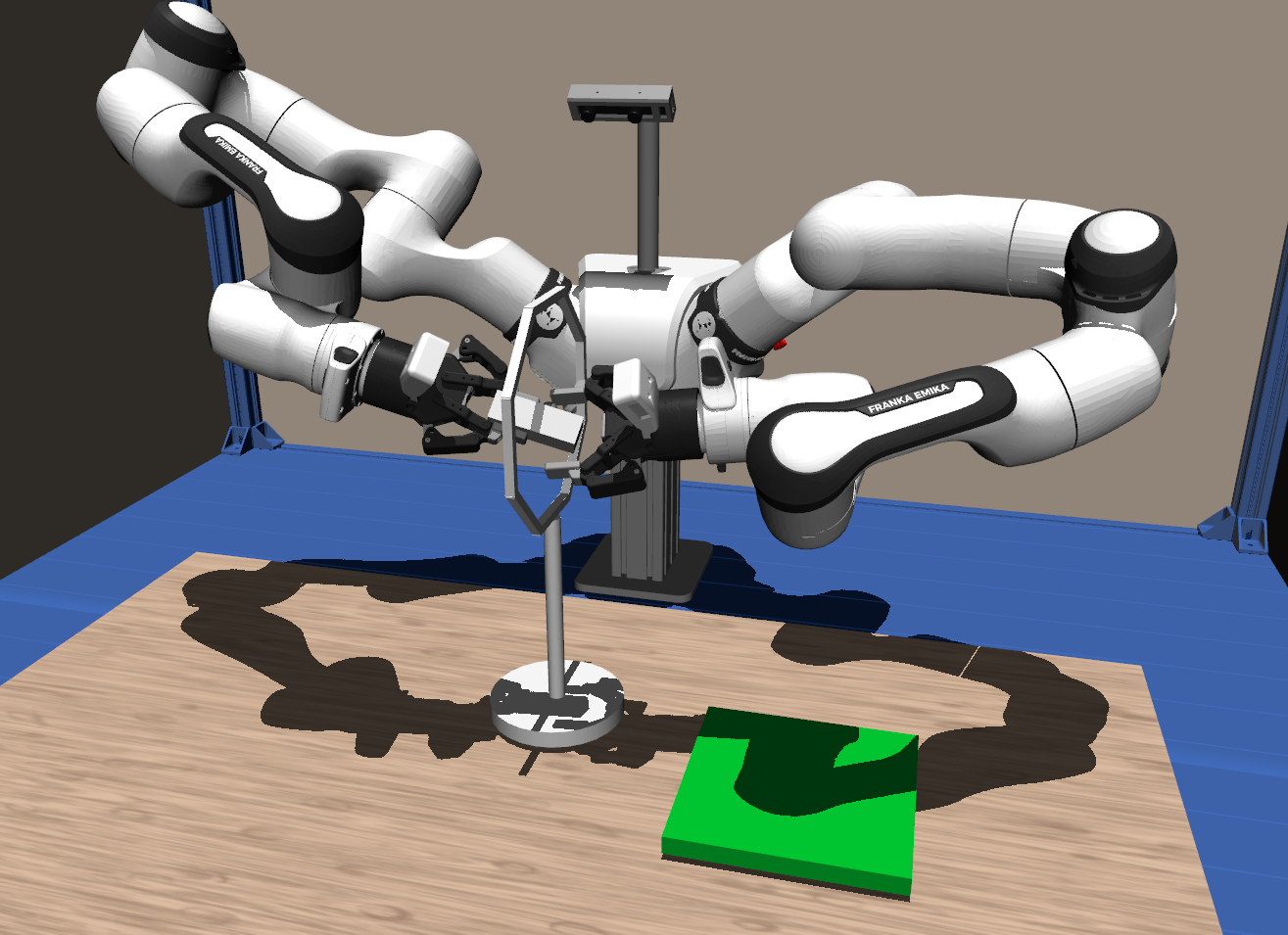} & \textbf{\transfergate} & Hand over a box between arms before placing it onto a mat. The box has to be passed through a gate. & Sequential Handoff & \checkmark& $\times$ & $523$ & pick up the box, pass the box through the ring, grab the box with the other hand, place the box on the mat & box, mat and goal positions, box and mat on each half of the table while the gate is in between\\

\includegraphics[width=3cm]{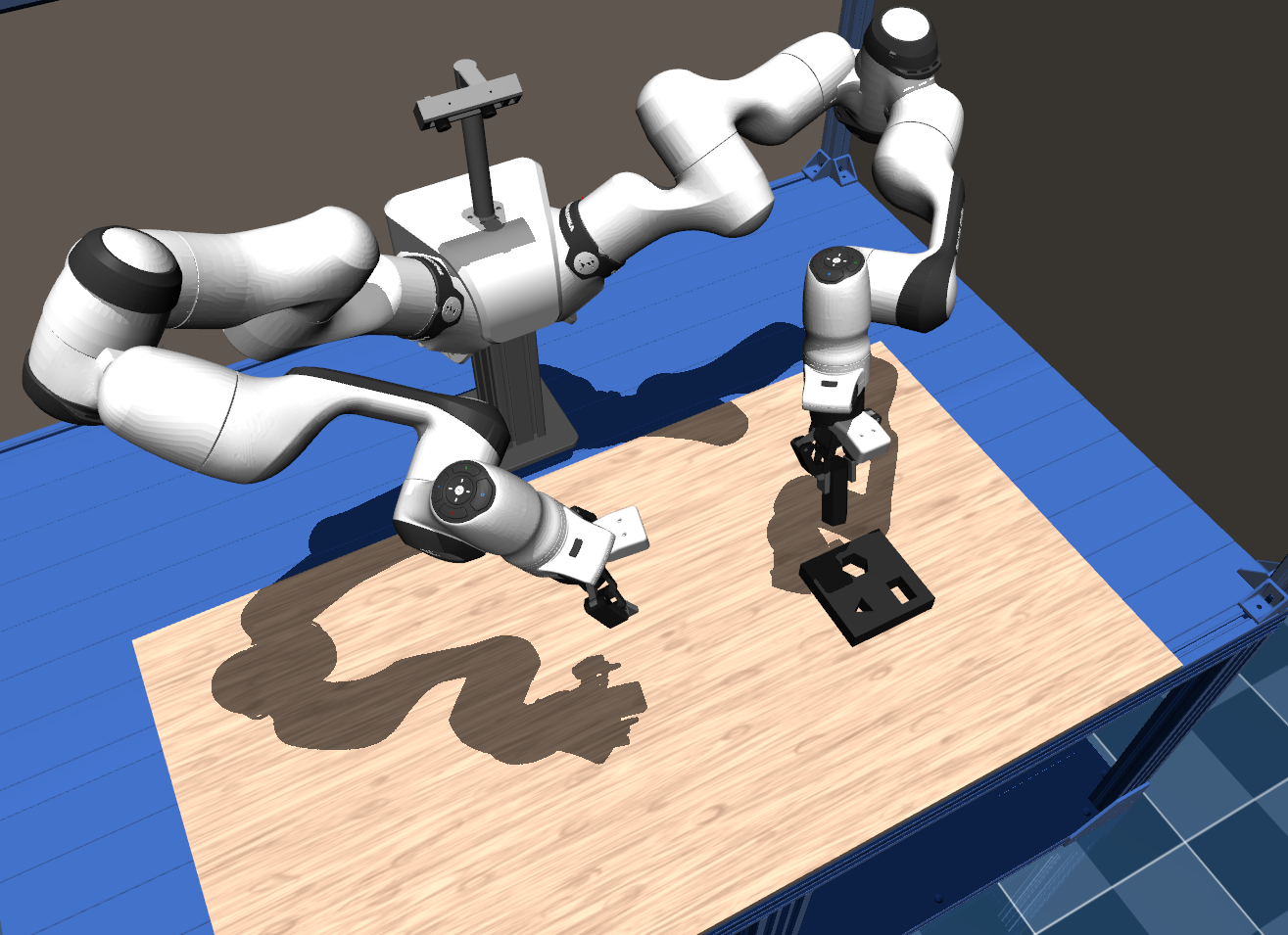} & \textbf{\transferreorient} & The right arm picks up a peg and hands it over to the left arm in such a way that the left arm is then able to insert it into a socket. & Sequential Handoff & \checkmark & $\times$ & $505$ & pick up the peg, bring both grippers into contact with the peg, transfer the peg to the other gripper, insert the peg into the matching socket & peg and socket positions, each on one half of the table \\

\includegraphics[width=3cm]{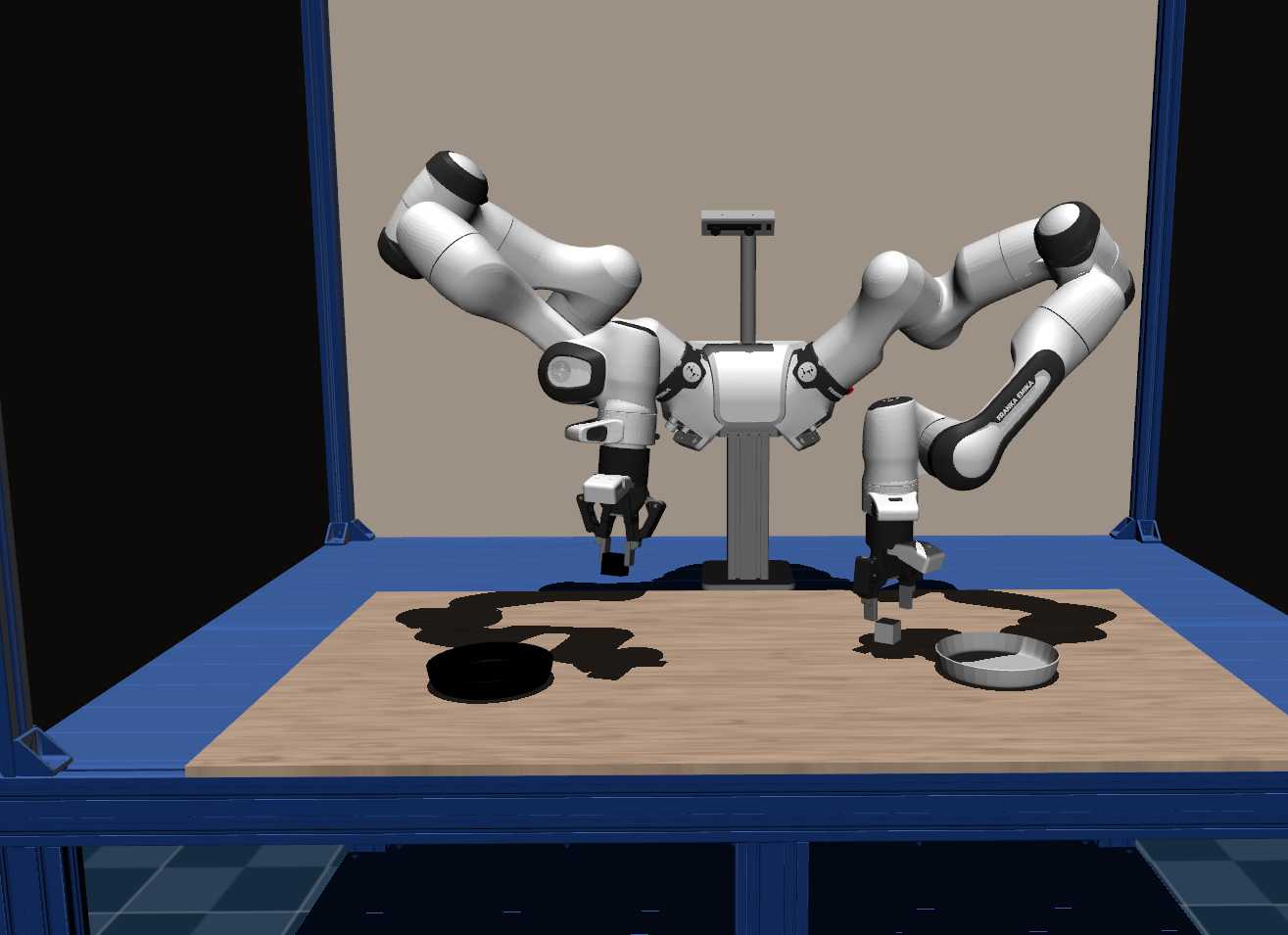} & \textbf{\binsort} & Sort two cubes into matching bowls, testing simultaneous execution rather than direct cooperation. & Parallel Execution & \checkmark & \checkmark & $216$ & pick up at least one cube, pick up the other cube or place the picked cube in the correct bowl, place at least one cube in the correct bowl, pick up the second cube, place the second cube in the correct bowl & cube and bowl positions \\

\bottomrule
\end{longtable}

\end{landscape}

\end{document}